\documentclass[journal]{IEEEtran}
\usepackage{overpic}
\usepackage{times}
\usepackage{epsfig}
\usepackage{graphicx}
\usepackage{subfigure}
\usepackage{amsmath}
\usepackage{amssymb}
\usepackage{url}
\usepackage{graphicx}
\usepackage{animate}
\usepackage{multirow}
\usepackage{color,xcolor}
\ifCLASSINFOpdf
\else
\fi

\usepackage[switch,columnwise]{lineno}

\begin{document}
	%
	\title{{Face Inverse Rendering via Hierarchical Decoupling}}
	%
	%
	
	\author{Meng Wang, Xiaojie Guo, Wenjing Dai, and Jiawan Zhang
	\thanks{Manuscript received July 18, 2021; revised June 24, 2022; accepted
			August 8, 2022. This work was supported by the National Natural Science Foundation of China under Grant 62072327 and Grant 62172295, and National Key Research and Development Program of China under 2019YFC1521200. The associate editor coordinating the review of this manuscript
			and approving it for publication was Prof. Cheung, Sen-Ching Samson. \emph{(Corresponding
			author: Xiaojie Guo.)}}
	\thanks{M. Wang, X. Guo, and J. Zhang are with the College of Intelligence and Computing, Tianjin University, Tianjin 300350, China  e-mail: (autohdr@gmail.com, xj.max.guo@gmail.com, jwzhang@tju.edu.cn). W. Dai is with the Department of Technology, Management and Economics Sustainability, Technical University of Denmark, Denmark e-mail:(weda@dtu.dk)}
	}
	
	%
	%

\markboth{Journal of \LaTeX\ Class Files,~Vol.~14, No.~8, August~2015}%
{Shell \MakeLowercase{\textit{et al.}}: Bare Demo of IEEEtran.cls for IEEE Journals}
%



\maketitle

\begin{abstract}
	Previous face inverse rendering methods often require synthetic data with ground truth and/or professional equipment like a lighting stage. However, a model trained on synthetic data or using pre-defined lighting priors is typically unable to generalize well for real-world situations, due to the gap between synthetic data/lighting priors and real data. Furthermore, for common users, the professional equipment and skill make the task expensive and complex. In this paper, we propose a deep learning framework to disentangle face images in the wild into their corresponding albedo, normal, and lighting components. Specifically, a decomposition network is built with a hierarchical subdivision strategy, which takes image pairs captured from arbitrary viewpoints as input. In this way, our approach can greatly mitigate the pressure from data preparation, and significantly broaden the applicability of face inverse rendering. Extensive experiments are conducted to demonstrate the efficacy of our design, and show its superior performance in face relighting over other state-of-the-art alternatives. { Our code is available at \url{https://github.com/AutoHDR/HD-Net.git}}.	
\end{abstract}


%
\IEEEpeerreviewmaketitle

\section{Introduction}
\IEEEPARstart{F}{ace} inverse rendering can be viewed as a task of disentangling human face images into their albedo maps and shading maps, while the latter ingredient can be further decomposed into two components,~\emph{i.e.}, normal and illumination. 
The problem of face inverse rendering is severely ill-posed in nature, because the number of unknowns
to be recovered is multiple times as many as that of the given measurements. A wide spectrum of applications could benefit from FIR, for instance, face relighting in virtual/augmented reality and style transfer, to name just a few. 
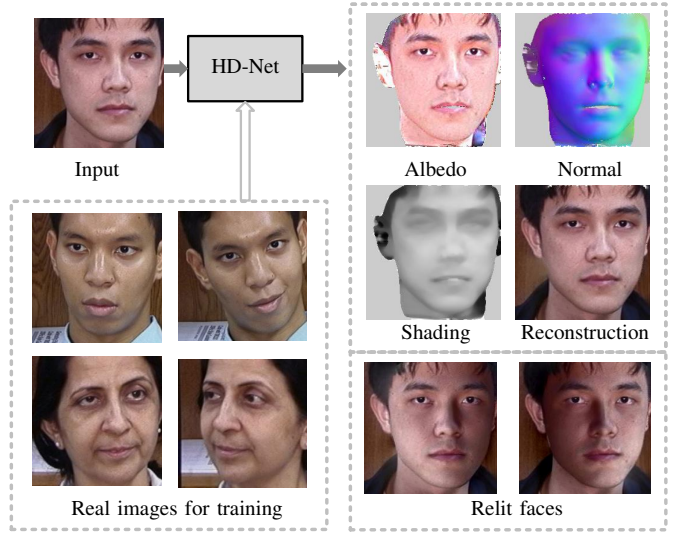
\begin{figure}[ht]
	\begin{center}
		\begin{overpic}[width=\linewidth]{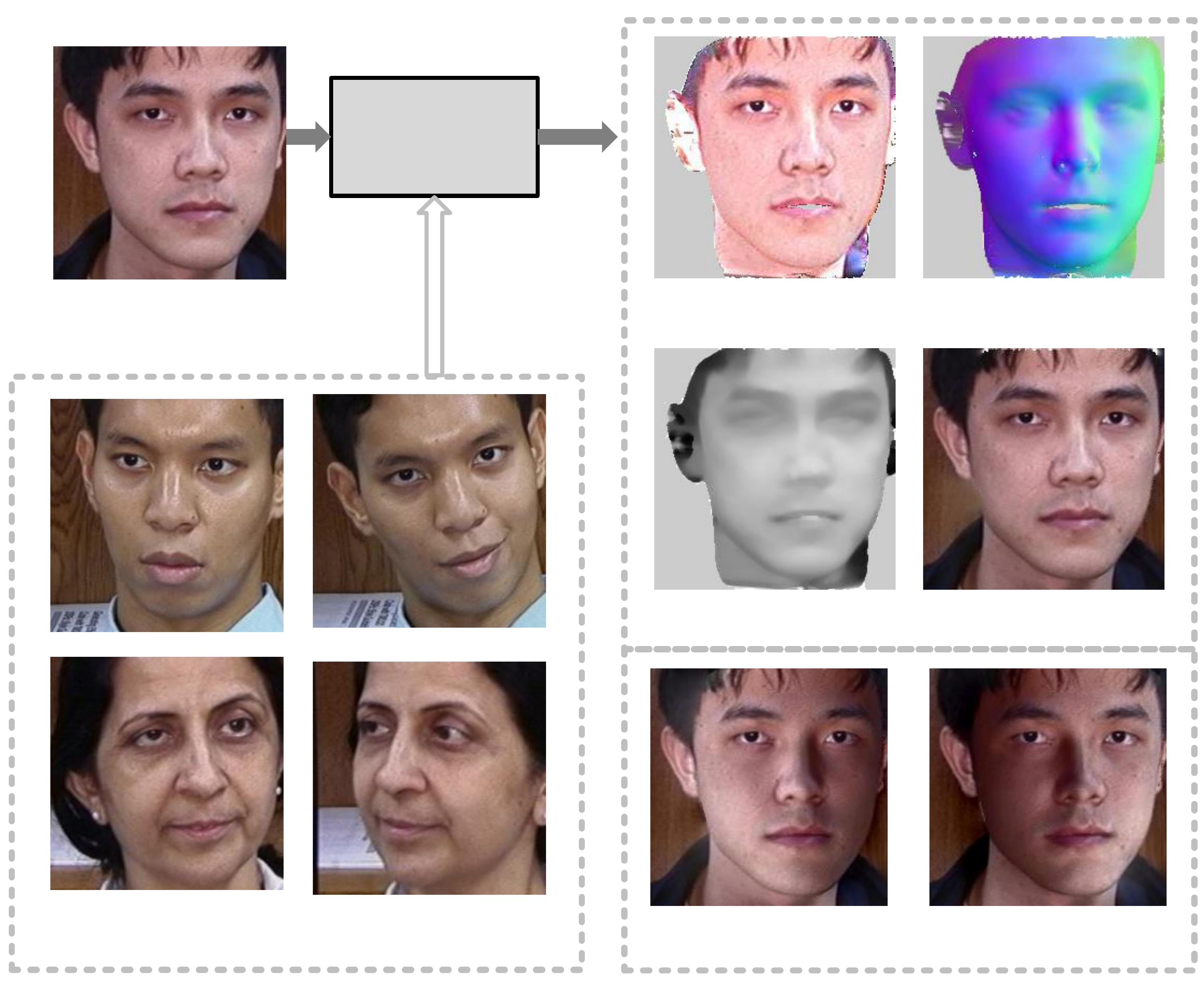}		
			\put(10.5, 54.5){\footnotesize {Input}}
			\put(31, 70){\footnotesize {HD-Net}}

			\put(60, 54.5){\footnotesize {Albedo}}
			\put(83, 54.5){\footnotesize {Normal}}
			\put(59.5, 30){\footnotesize {Shading}}
			\put(77.5, 30){\footnotesize {Reconstruction}}
			
			\put(10,3.5){\footnotesize {Real images for training}}
			\put(70,3.5){\footnotesize {Relit faces}}
			
		\end{overpic}
	\end{center}
	{\caption{Illustration of our proposed \emph{HD-Net}, which decomposes real face images into several components. The faces can be relit through changing the lighting conditions.}}
	\label{process_img}
\end{figure}

To ease the ill-posedness, one technical line, with ~\cite{debevec2000acquiring,sun2019single,lattas2020avatarme,smith2020morphable} as representatives, seeks help from physical equipment on acquiring face images from different viewpoints and under varying lighting conditions. Though these methods can mitigate the difficulty of decomposition by providing ground-truth information, they demand professional photographing skills/tools and complicated preparation under well-controlled circumstances, significantly limiting the applicability to typical users. For the sake of releasing the professional requirement, several algorithms,~\emph{e.g.}~\cite{sengupta2018sfsnet} and~\cite{egger2018occlusion}, have been designed to learn face components (partially) on synthetic data. Despite the improvement, these methods still suffer from the gap left by the relatively simple distribution of synthetic data to real scenarios. 
Concerning the drawbacks of the aforementioned methods, it is highly desirable to design an equipment-free and real data-fitted model, which is the goal of this work. However, three main challenges impede moving towards the desire as follows:
\begin{enumerate}
	\item \textbf{\textit{Severe ill-posedness}}. Separating an image in the wild into several components (one-to-three decomposition task under the Lambertian model in this work) is in nature heavily under-determined.
	\item \textbf{\textit{High variation}}. The faces often appear to be loosely controlled. Pose and expression affect the appearance of faces in images, which, compared to global 2D changes, will result in higher complexity of the decomposition even in a supervised fashion.
	\item \textbf{\textit{Unavailable ground-truth}}. In practice, it is expensive and complicated, if not impossible, to capture ground-truth information for respective components of real face images. Thus, solving the target problem without supervision seems to be extremely hard.
\end{enumerate}

\subsection{Consideration \& Contribution}

\begin{figure*}
	\begin{center}
		\centering	
		\begin{overpic}[width=\linewidth]{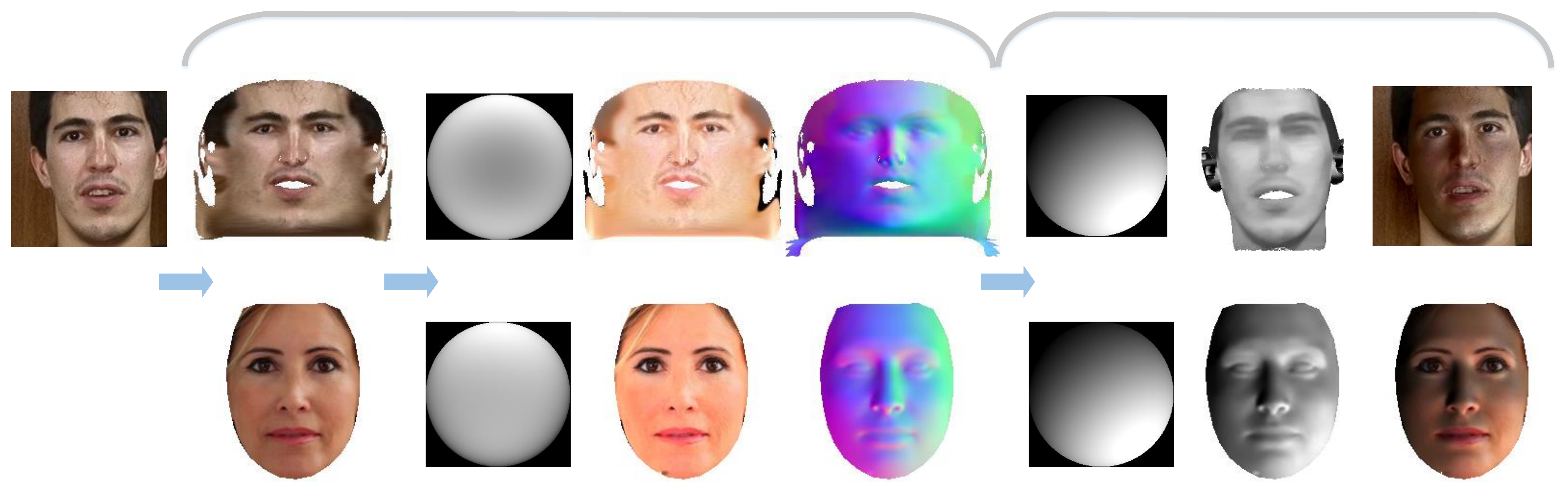}		
			\put(30,27.5){\small \color{black}{\textbf{Inference phase}}}
			\put(79,27.5){\small\color{black}{\textbf{Relighting}}}		
			
			\put(11, 9.5){\footnotesize \color{black}\rotatebox{90}{\textbf{Unwarping}}}
			\put(25.5,9.5){\footnotesize \color{black}\rotatebox{90}{\textbf{Decomposing}} }
			\put(63.5,11){\footnotesize \color{black}\rotatebox{90}{\textbf{Warping}} }
			
			\put(5,-1.5){\footnotesize \color{black}{Input}}		
			\put(17,-1.5){\footnotesize \color{black}{Texture}}			
			\put(31,-1.5){\footnotesize \color{black}{SH}}
			\put(41,-1.5){\footnotesize \color{black}{Albedo}}
			\put(54,-1.5){\footnotesize \color{black}{Normal}}	
			\put(67.5,-1.5){\footnotesize \color{black}{Target SH}}
			\put(79.5,-1.5){\footnotesize \color{black}{Shading}}
			\put(92,-1.5){\footnotesize \color{black}{Relit}}			
		\end{overpic}
		\caption{{The procedure for our face inverse rendering and relighting.} The input is a single unwarped texture/image, which is then decomposed into lighting (or spherical harmonics, SH), albedo, and normal. For relighting, the decomposed SH lighting is replaced by a target SH provided by users. The shading map is constructed by recomposing the normal and target SH, and the relit shading further drags the decomposed albedo into a newly generated image. The first row tests on non-aligned face image datasets GT~\cite{gt2007data} with the help of unwarping and warping, and the second row tests on FFHQ~\cite{karras2019style}. }		
		\label{figure1_top}
		
	\end{center}
\end{figure*}
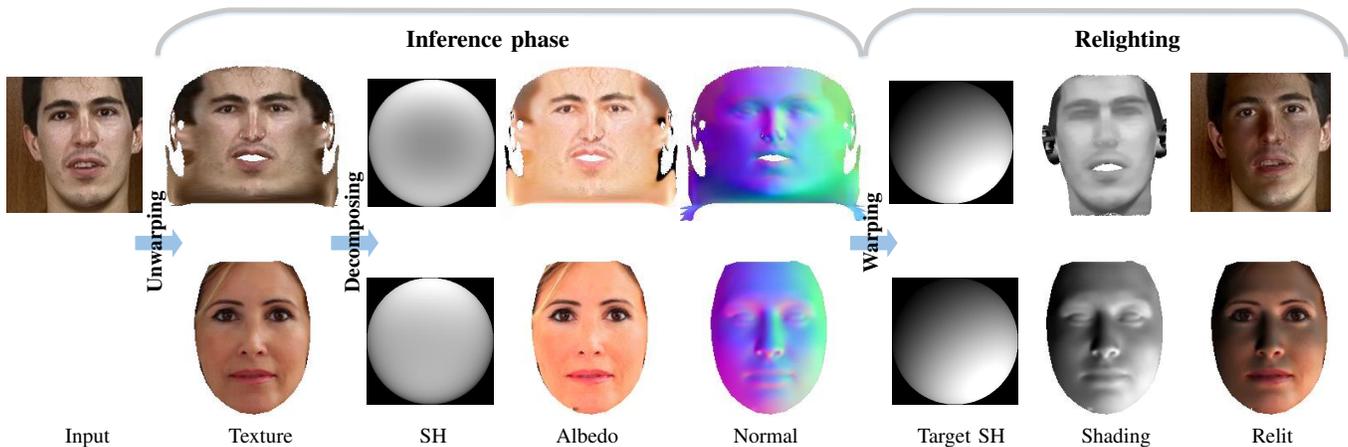

Regarding the ill-posed characteristic, a one-to-three decomposition problem, like mapping a face image into its albedo, normal and lighting components (the target of this work, please see Figure~\ref{figure1_top}), is technically much more difficult than a one-to-two problem, \emph{e.g.} separating a face into its albedo and shading maps (\emph{a.k.a.} intrinsic image decomposition), because the searching space exponentially expands as the number of unknowns to recover increases. To possibly relieve the difficulty, converting a one-to-three decomposition problem into two decoupled one-to-two sub-problems could be a good choice. Fortunately, under the Retinex theory~\cite{land1971lightness} and the assumption of Lambertian reflectance~\cite{jacobs2005lambertian}, the albedo can play such a pivotal role in decoupling the original problem. Driven by this fact, we design a hierarchical strategy to achieve the goal. More concretely, a face image is firstly disassembled into its albedo and shading maps, then the shading is further decoupled into two elements, \emph{i.e.} normal and lighting. 

As for the high variation, thanks to the strong structure of faces, aligning different faces to a canonical status would effectively alleviate the issue. In the literature, a number of face alignment techniques have been proposed, among which the 3D Morphable Model (3DMM) and its follow-ups~\cite{tran2018nonlinear,tran2019learning} are arguably the most representative. Even with the above two points being properly disposed, another obstacle is that no ground-truth information is available to guide the training procedure. In other words, effective constraints need to be imposed on the desired solutions. We assume that, after alignment, the albedo and normal maps of the same person should be closely similar (consistency). In addition, the shading map, although it might be diverse for different face images, should be largely smooth.

Based on the above consideration, we customize a deep network to hierarchically decompose face images into three components, including albedo, normal and illumination/lighting. The main contributions of this work can be summarized as follows: 
\begin{enumerate}
	\item For tackling the one-to-three face decomposition, we propose a hierarchical strategy to alternatively solve two one-to-two smaller problems, which significantly reduces the complexity of original problem.
	\item We employ the 3D face alignment technique to deal with the high variation of face appearance in terms of pose and expression, which further shrinks the freedom-degree of target space.  
	
	\item {In an unconstrained setting, simple yet effective constraints, such as the albedo and normal consistency on aligned faces of the same person as well as the piece-wise smoothness on the shading map, are exploited to make the problem tractable.}
\end{enumerate}
Extensive experiments are conducted to reveal the effectiveness of our design, and show its superiority over other state-of-the-art methods both quantitatively and qualitatively.

\section{Related Work}
A variety of inverse rendering methods have been devised over the last decades. This section briefly reviews classic and contemporary techniques closely related to this work. \\
%

\textbf{Equipment-based solutions:} Debevec~\emph{et al.}~\cite{debevec2000acquiring} and Sun~\emph{et al.}~\cite{sun2019single} employed specialized light stages to capture reflectance information of human faces. Weyrich~\emph{et al.}~\cite{weyrich2006analysis,weyrich2005measurement} introduced a face-scanning dome with 16 cameras and 150 light sources to capture sequence with two different exposure settings. Ghosh~\emph{et al.}~\cite{ghosh2011multiview}, Wang~\emph{et al.}~\cite{wang2020single} and Lattas~\emph{et al.}\cite{lattas2020avatarme} built a setup for multi-view face scanning with several cameras and light sources. Utilized a consistent environment or a specific device to capture images for estimating face appearance has been proposed in~\cite{saito2017photorealistic, yamaguchi2018high, sun2019single, nestmeyer2020learning}. Though effective, the data they used is captured under a specific environment with professional equipment, and these techniques rely on complicated systems and computing, which can hardly be used for consumer-level usage.

Instead of following the physical equipment-based pipeline, a couple of works benefit from the generalized bas-relief (GBR) transformation (called photometric stereo). The early attempt via shape and surface reflectance decomposition based on the photometric stereo goes to  ~\cite{woodham1980photometric}. This manner has been widely applied, especially in normal recovery, such as~\cite{chen2002recovery,barsky20034,basri2007photometric,ikeda2008color,shi2010self,hauagge2015photometric,chakrabarti2016single}. Ogun~\emph{et al.} \cite{ogun2010determination} proposed a method to seek the surface reflectance without computing surface gradients. Shi~\emph{et al}. \cite{shi2010self} developed a complete auto-calibration approach for estimating surface normals and albedos. Hauagge~\emph{et al.} \cite{hauagge2015photometric} computed a per-pixel statistic over a stack of images, and combined local geometry at each point with illumination to decompose ambient occlusion, albedo and illumination. In~\cite{chakrabarti2016single}, the authors applied color filters in front of 3 LED lights and estimated the surface normal under the constraint of piece-wise smooth albedo. 

Recently, Doris~\emph{et al.} \cite{antensteiner2019single} proposed a network trained with 3 differently colored illuminations to estimate albedos and normals from a single shot image. Hashimoto~\emph{et al.} \cite{hashimoto2019uncalibrated} assumed the albedo is constant except for edges and calculated albedo and normal solely from the image sequence. In~\cite{hamaen2020multispectral}, a set-up with different wavelengths of light sources is used to capture images, and the multispectral photometric stereo and intrinsic image decomposition are adopted to solve the ambiguity of albedo and light. They can achieve the relighting effects, however, the goal of GBR is more like to estimate accurate normal. Besides, surface reflectance is often considered as an intermediate product. Besides, all methods based on photometric stereo need to face the camera and take images from a fixed viewpoint. Our proposed method is not restricted by viewpoints and poses.\\

\textbf{Supervised and unsupervised solutions: } Collecting large-scale decomposition datasets with fine labels from the real world is considerable expensive for supervised inverse rendering methods. Alternatively, several works use synthetic data, such as IIW dataset~\cite{li2018cgintrinsics} for intrinsic image decomposition of indoor scenes and synthetic face dataset~\cite{sengupta2018sfsnet} for face inverse rendering based on the physical-based model, to do the job. In~\cite{janner2017self}, a self-supervised method is customized for intrinsic image decomposition with a pre-trained shading model trained with synthetic data. Soumyadip~\emph{et al.} \cite{sengupta2018sfsnet} learned low frequency from synthetic data. Meanwhile, high-frequency details in real data are captured using shading cues from 'pseudo-supervision'. The reconstruction loss to real data with the pseudo label would not be backward propagated accurately. Besides, the models trained on synthetic data often cannot work well on real face images.

As for unsupervised learning, several works explore the priors of individual components, and build shared-weight networks to decompose each component from multiple images. Lettry~\emph{et al.} \cite{lettry2018unsupervised} proposed an intrinsic image decomposition network to extract two local features by using the same weights with the help of unlabeled time-varying sequence images. Ma~\emph{et al.} \cite{ma2018single} designed a network that requires neither ground truth nor priors. They drew connections between single image-based methods and multi-image-based approaches to show that one can benefit from the other. Besides, they presented a two-stream convolution neural network, which takes a pair of varying illumination images as input. Even though the models of~\cite{lettry2018unsupervised} and~\cite{ma2018single} are powerful, they only fit for intrinsic image decomposition, thus cannot be used for lighting transfer on account of the constraint ability of intrinsic decomposition. Michael~\emph{et al.} \cite{janner2017self} proposed a network with a shared convolutional encoder and three decoders for reflectance, shape, and lighting, respectively. However, they need the trained model with ground truth as an initialization. With the help of each component's specific physical meaning, Shu~\emph{et al.}\cite{shu2017neural} introduced a weakly unsupervised network by adding additional constraints. However, the ambiguity in the magnitude of lighting leads to unrealistic results. Besides, their results also lose high-frequency details, resulting in poor performance. Recently, some works can produce face decomposition individuals as intermediate components, such as~\cite{tewari2017mofa, tewari2019fml, tran2019learning, gecer2019ganfit}. In contrast, our approach has a targeted approach to face inverse rendering in the wild in an unsupervised manner, obtaining high-frequency components via a hierarchical decoupling network.

\begin{figure*}[t]
	\begin{center}
		\begin{overpic}[width=\linewidth]{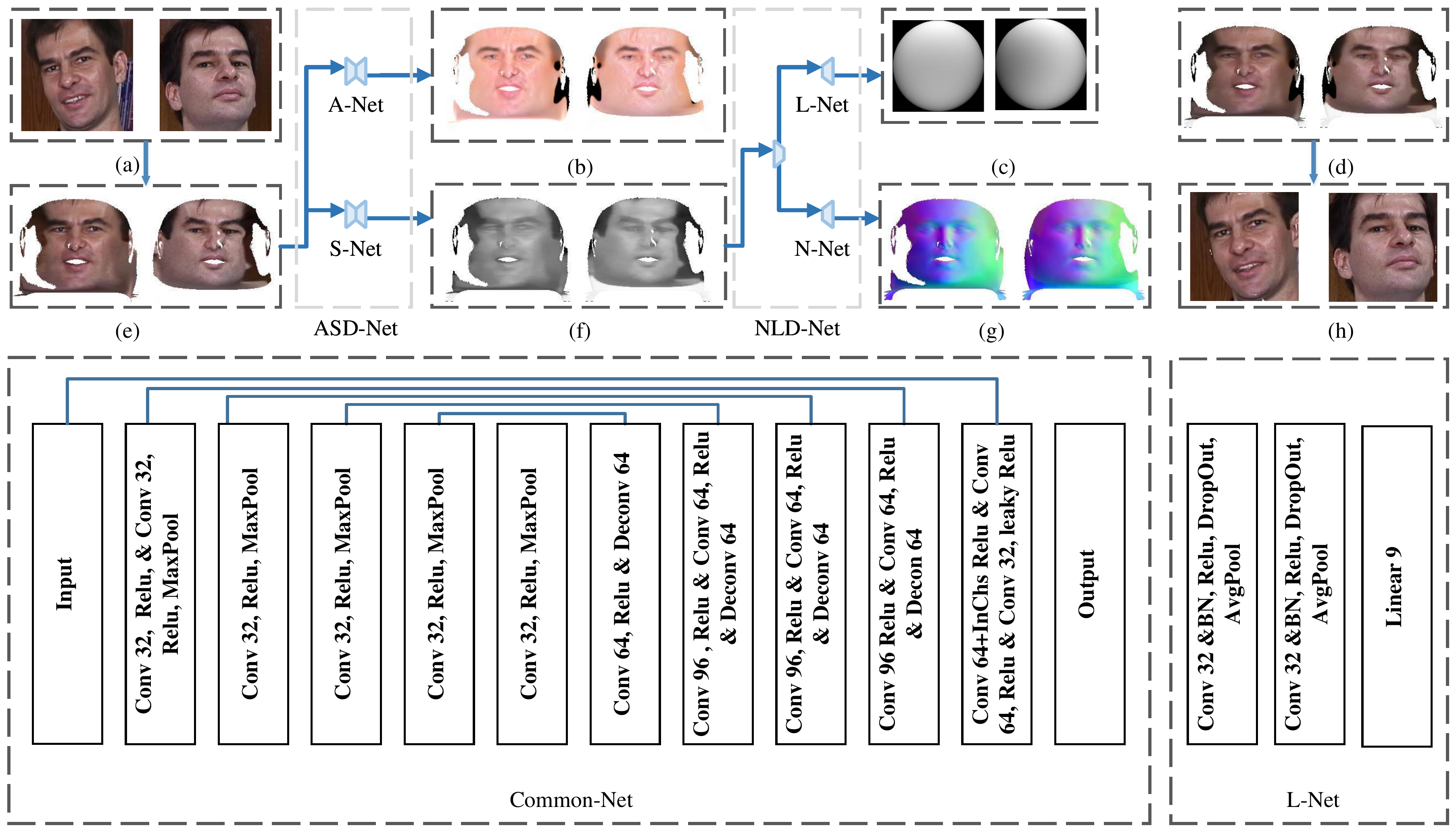}
			
			\put(10,4.2){$E_{c1}$}
			\put(16.3,4.2){$E_{c2}$}
			\put(23,4.2){$E_{c3}$}
			\put(29.5,4.2){$E_{c4}$}
			\put(35.7,4.2){$E_{c5}$}
			\put(42,4.2){$D_{c1}$}
			\put(48.2,4.2){$D_{c2}$}
			\put(54.5,4.2){$D_{c3}$}
			\put(61,4.2){$D_{c4}$}
			\put(67.5,4.2){$D_{c5}$}
			
			\put(83,4.2){$D_{l6}$}
			\put(89,4.2){$D_{l7}$}
			\put(95,4.2){$D_{l8}$}
			
		\end{overpic}
		\caption{{Overview of our hierarchical decoupling network (\emph{HD-Net})}. The network consists of two decoupling sub-nets, \emph{i.e.} the albedo and shading decoupling network (\emph{ASD-Net}) and the normal and lighting decoupling (\emph{NLD-Net}). First, the input images are unwarped, from (a) to (e). \emph{ASD-Net} decomposes the unwarped textures (e) into albedo (b) and shading (f). \emph{NLD-Net} disentangles shading (f) into normal (g) and light (c). Finally, the reconstructed unwarped textures are warped back from (d) to (h). 
		}
		\label{figure2_network}
	\end{center}
\end{figure*}
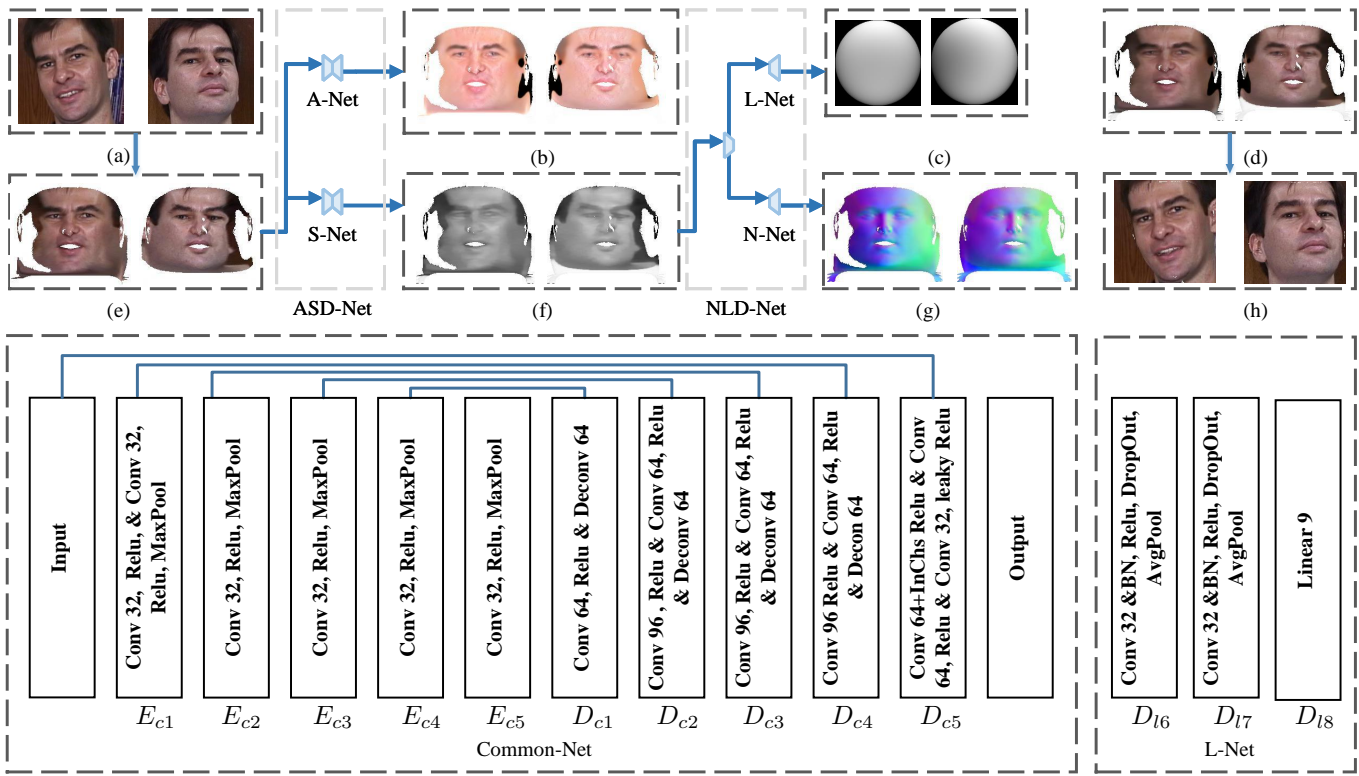

\section{Hierarchical Decoupling Network}

Formally, we denote the input image, albedo, normal, and shading at position $p$ as \emph{I(p)}, \emph{A(p)}, \emph{N(p)}, and \emph{S(p)}, respectively; \emph{L} is represented as distant lighting. Then the rendering under Lambertian reflectance~\cite{jacobs2005lambertian} can be formulated as follows:

\begin{equation}
	I(p) = \mathcal{F}_{render}(A(p),N(p),L),
	\label{con:anl}
\end{equation}
where $\mathcal{F}_{render}$ is a physical-based rendering function for reconstructing the input image \emph{I(p)}. For intrinsic image decomposition, the reconstruction is typically written as:
\begin{equation}
	I(p) = \mathcal{F}_{recons}(A(p),S(p)) = A(p) \cdot S(p),
	\label{con:as}
\end{equation}
where the operator $\cdot$ designates the element-wise product. From the above two formulations, it is clear to see that the albedo can be viewed as a bridge between intrinsic decomposition and physical-based inverse rendering.

Following previous works~\cite{wang2008face,shu2017neural,sengupta2018sfsnet,zhou2019deep}, for position $p$, the normal is $N(p)=[x_p,y_p,z_p]^T$, the lighting $L$ can be expressed as a 9-dimensional spherical harmonics coefficient $l=[l^{1},l^{2},...,l^{9}]^T$. Accordingly, the spherical harmonic basis $h_p = [h_p^{1},h_p^{2},...,h_p^{9}]^T$ can be represented as:
\begin{small} 
	\begin{align*}
		h_p^{1}&=\frac{1}{\sqrt{4 \pi}},  & h_p^{2}&=\sqrt{\frac{3}{4 \pi}} y_p, &	h_p^{3}&=\sqrt{\frac{3}{4 \pi}} z_p,\\
		h_p^{4}&=\sqrt{\frac{3}{4 \pi}} x_p,  &h_p^{5}&=3 \sqrt{\frac{5}{12 \pi}} x_p y_p,  &\\ 
		h_p^{6}&=3 \sqrt{\frac{5}{12 \pi}} y_p z_p, & h_p^{7}&=\frac{1}{2} \sqrt{\frac{5}{4 \pi}}\left(3 z_p^{2}-1\right),  &  \\
		h_p^{8}&=3 \sqrt{\frac{5}{12 \pi}} x_p z_p, &h_p^{9}&=\frac{3}{2} \sqrt{\frac{5}{12 \pi}}\left(x_p^{2}-y_p^{2}\right). & 
	\end{align*}
\end{small}
On the one hand, the shading $S_p$ can be reconstructed from the normal $N(p)$ and lighting $L$, represented as: 
\begin{equation}
	S(p)=\mathcal{F}_{shading}(N(p), L) = h_p^T  l.
	\label{con:nls}
\end{equation}
On the other hand, the lighting information can be simply computed from the predicted normal and shading maps via least square optimization.



\subsection{Network architecture}




As discussed above, we regard the face inverse rendering as a recursive decoupling problem, which converts a one-to-three decomposition into two smaller/simpler one-to-two tasks. It is easy to find refined patterns with a simple, straightforward solution until a certain level of simplicity is achieved. The hierarchical decomposition can be independently applied to selected subsets in a divide and conquer manner, significantly reducing the overall computation cost and the difficulty of adjusting weights in the loss function during training. We emphasize that the goal of this paper is to learn a model that decomposes the unconstrained human face images into three components, namely, albedo maps, normal maps, and lightings. The overall architecture is illustrated in Figure~\ref{figure2_network}. 

The pivot of our method is the albedo, which is the bridge between physical-based inverse rendering and intrinsic image decomposition. Therefore, we take advantage of the relationship and build a connection between decoupling networks. Specifically, as shown in Figure~\ref{figure2_network}, the input image $I$ is firstly decomposed into the albedo $A$ and shading $S$ via the albedo and shading decoupling network, denoted as \emph{ASD-Net}, which contains two branches including \emph{A-Net} and \emph{S-Net}. The shading $S$ can be then decoupled into the normal $N$ and light $L$ by the normal and lighting decoupling network (\emph{NLD-Net}), which also contains two branches, called \emph{N-Net} and \emph{L-Net}. The following details the architectures of the two sub-networks, say \emph{{ASD-Net}} and \emph{{NLD-Net}}. 

The sub-networks of \emph{A-Net}, \emph{S-Net} and \emph{N-Net}, except for the \emph{L-Net} are of the same architecture, called \emph{Common-Net}. {The \emph{L-Net} and \emph{N-Net} share the same encoder in \emph{NLD-Net}, the encoded features $E_{c5}$ at the 5th layer of \emph{NLD-Net} are fed into \emph{L-Net} for lighting prediction. We employ the classical U-shaped based network from Noise2Noise~\cite{lehtinen2018noise2noise} as our Common-Net. The main reason is to largely exclude other possible influences from sophisticated network architectures and focus on our proposed strategy. Even with such a simple net, our performance is promising, which reveals the effectiveness of our design and verifies the main claims.}

\noindent\textbf{Albedo and shading decoupling network.}
Given an input pair of images, we firstly get the unwarped texture as the input of our~\emph{{ASD-Net}}. As illustrated in Figure~\ref{figure2_network}, The~\emph{{ASD-Net}} takes an unwarped texture of image pair as input and gradually learns the decoupled parameters. \emph{{ASD-Net}} has two branches,~\emph{A-Net} and~\emph{S-Net} predicting albedo and shading, respectively. The outputs are albedo map and shading map. Our subdivision process follows a simple update rule, directed by the intrinsic decomposition model as in Eqn.~\eqref{con:as}.\\

\noindent\textbf{Normal and lighting decoupling network.}
Based on the shading generated from the first decoupling network~\emph{ASD-Net}, we can further apply a subdivision strategy for shading by a second decoupling network \emph{NLD-Net}, which is to decompose the shading map into normal map and lighting. It also has two sub-networks,~\emph{N-Net} and~\emph{L-Net} for normal and lighting prediction, respectively.\\

\noindent Tuning parameters could be sensitive and difficult without ground truth. Our hierarchical decoupling network employs a step-wise subdivision strategy to reduce the sensitivity of parameter tuning and facilitate the rapid identification of suitable decomposition parameters.

\subsection{Loss design}


As we do not have the ground-truth components of the input image, additional constraints need to be imposed to guide our hierarchical network. We can utilize the properties of individual components. Please notice that to avoid the disturbance from various expressions and poses of faces, we apply~\cite{tran2018nonlinear} to unwarp the face images for alignment.\\

\noindent\textbf{Albedo and shading losses.} 
For a certain person, the albedo should be closely similar or consistent. Thus, the consistency of albedo in paired images can offer a piece of information to constrain albedo learning. While shading is partially influenced by illumination, which should be piece-wise smooth~\cite{zhang2019kindling}. Considering these two aspects, we introduce the shading smoothness loss in the gradient domain, and the consistent loss on albedo to \emph{ASD-Net} for albedo (\emph{A-Net}) and shading (\emph{S-Net}) prediction, respectively. The loss function $\mathcal{L}_{ASD}$ can be expressed as follows:
\begin{equation}
	\mathcal{L}_{ASD} = \lambda_{a}\mathcal{L}_{a} + \lambda_{s}\mathcal{L}_{s}.
	\label{con:albedo}
\end{equation}
Concretely, $\mathcal{L}_{a}$ is defined as $\|{ A_{i}}-{A_{j}}\|_1$, where $\|\cdot\|_1$ means the $\ell_1$ norm. This term regularizes the fidelity between the estimated albedos ${A_{i}}$ and ${A_{j}}$  from paired images $I_i$ and $I_j$. 
The shading map should be piece-wise smooth under a distant light. Similar to~\cite{zhang2019kindling}, $\mathcal{L}_{s}$ is defined as $\lambda_{s}(\|\frac{\nabla {S_{i}}}{max(|\nabla {S_{i}}|,\xi)}\|_1
+ \|\frac{\nabla {S_{j}}}{max(|\nabla {S_{j}}|,\xi)}\|_1
)$, where $\nabla {S}$ stands for the derivative operator of $\nabla {S_{x}}$ and $\nabla {S_{y}}$ in the first order on shading predicted from pair of image, and $\xi$ is a small positive constant (0.01 in this work) for avoiding zero denominator. The non-negative coefficients $\lambda_a$ and $\lambda_s$ balance the importance of the corresponding terms.\\ 

\noindent\textbf{Normal and lighting losses.} Suppose that we have already had estimated albedo and shading from the \emph{ASD-Net}, it is still hard to disentangle the normal and light due to the ambiguity between them. A number of strategies have been devised for normal estimation, such as~\cite{blanz1999morphable,zhu2017face,tran2018nonlinear}. To ease the normal estimation and the lighting, we introduce the estimated normal by 3D Morphable Model~\cite{blanz1999morphable} as an initialization, which acts as a reference for our estimation.
The loss function $\mathcal{L}_{NLD}$ is as follows:
\begin{equation}
	\mathcal{L}_{NLD} = \lambda_{n}\mathcal{L}_{n} +  \lambda_{l}\mathcal{L}_{l},
	\label{con:nl}
\end{equation}
{where $\mathcal{L}_{n}$ is defined as $\| \bar{N}_i-N_i\|_2^2 + \| \bar{N}_j-N_j\|_2^2$ with $\bar{N}_{i}$ and $\bar{N}_{j}$ initialized normal maps for coarse training, $N_{i}$ and $N_{j}$ are the predicted normal maps from \emph{N-Net}, and $\|\cdot\|_2$ stands for the $\ell_2$ norm.} Moreover, $\mathcal{L}_{l}$ adopts $\|l_{i}-\hat{l}_{i}\| _1 + \| l_{j}-\hat{l}_{j}\|_1$ for regularizing the lighting component. Please note that, under the Lambertian shading model, lights can be represented by 9-dimensional spherical harmonic coefficient vectors. In the term $\mathcal{L}_{l}$, $\hat{l}_{i}$ and $\hat{l}_{j}$ are computed from the predicted shading maps and initialized normal maps using least square optimization, and $l_{i}$ and $l_{j}$ are predicted lighting from \emph{L-Net}. \\


\noindent\textbf{Reconstruction and adversarial losses.} The reconstruction loss function $\mathcal{L}_{rec}$ contains two terms. One is about image reconstruction via $\mathcal{L}_{Irec}$, and the other takes care of shading reconstruction by $\mathcal{L}_{Srec}$, which can be represented as follows:
\begin{equation}
\mathcal{L}_{rec} = \lambda_{Irec}\mathcal{L}_{Irec} + \lambda_{Srec} \mathcal{L}_{Srec} + \lambda_{adv} \mathcal{L}_{adv}.
\label{con:recon}
\end{equation}
The image reconstruction loss $\mathcal{L}_{Irec}$ is under the intrinsic decomposition model, while the shading reconstruction loss $\mathcal{L}_{Srec}$ is based on the physical-based shading reconstruction computed from the predicted normal and lighting. {More specifically, $\mathcal{L}_{Irec} = \| I_{i}-{A}_{i}\cdot{S}_i\|_1 + \| I_{j}-{A}_{j}\cdot{S}_j\|_1$, where $I_{i}$ and $I_{j}$ are paired input images, $A_{i}$ and $A_{j}$, and $S_{i}$ and $S_{j}$ are the albedo maps and shading maps predicted from \emph{ASD-Net}.} In addition, $\mathcal{L}_{Srec}$ is given as $\|{S_{i}}-\hat{S}_{i}\|_1 + \|{S_{j}}-\hat{S}_{j}\|_1$ with $\hat{S}_{i}$ and $\hat{S}_{j}$ are the shadings reconstructed by the predicted normal and lighting as in Eqn.~\eqref{con:nls}. In this way, along with the reconstruction of the input, the shading reconstruction loss minimizes the gap between intrinsic image decomposition and physical-based rendering. 
Besides, the discriminator is frequently used in unsupervised learning to distinguish which ones are real from fake ones. In our hierarchical decoupling network, we introduce an adversarial loss $\mathcal{L}_{adv}$ as in~\cite{shrivastava2017learning} for the reconstruction under the Lambertian model, which can further guarantee the reasonable reconstruction. 




\subsection{Training strategy}
\label{TrainingStragy}
To learn each component, we introduce a rough initialization and subdivision strategy into our hierarchical network. This has the potential to improve the accuracy of predicted components significantly. In other words, we divide the training process into two stages.\\

\noindent\textbf{Training stage 1.} The parameters are randomly initialized at the start of the training phase; it is not possible to decompose individual components without normal initialization. For this reason, we introduce coarse normals for initialization in order to guide the normal learning; otherwise, the initialization normals also conduct the light predictions with the help of predicted shading. \\

\noindent\textbf{Training stage 2.} {When the training of the network is converged in stage 1, our model is considered to be able to decompose the albedo, shading, normal, and lighting, effectively. However, due to the effects of coarse initialization, some errors will be backpropagated (\emph{e.g.}, the loss propagation between the predicted light and the light computed from the initial normal maps and predicted shading maps via least-square optimization). As a result, the model training might be inaccurate. Thus, a few adjustments are made to refine the result, including 1) we fix the parameters of \emph{A-Net}, \emph{S-Net} and \emph{N-Net} except for the decoder of \emph{L-Net}; 2) we replace the initialization normal with the predicted normal to refine light until convergence; and 3) we train our network with a lower learning rate without fixing parameters. After several iterations, our network can gradually close the gap.}



\subsection{Implementation details}

The entire hierarchical decoupling network is schematically illustrated in Figure~\ref {figure2_network}, which contains two sub-networks \emph{ASD-Net} and \emph{NLD-Net}. Specifically, the input is first passed through a $32\times 3\times 3$ convolutional layer without pooling, and then the obtained features are fed into \emph{A-Net} and \emph{S-Net} for albedo and shading prediction, respectively. Following that, the predicted shading is employed by \emph{N-Net} and \emph{L-Net} to predict normal and lighting. 
The network is trained on images with $256 \times 256$ size.
For the GT dataset~\cite{gt2007data}, we use a mask to expand the unwarped pair of images to $256\times256$ for training and testing, due to the aligned face images generated from unwarped function~\cite{tran2018nonlinear} is $192*224$. {In the training phase, we set $\lambda_{a}=0.25$, $\lambda_{s}=0.1$, $\lambda_{n}=0.5$, $\lambda_{l}=0.01$, $\lambda_{Irec}=0.25$ and $\lambda_{Srec}=0.01$. Besides, an adversarial loss with $\lambda_{adv}=0.001$ of the input face is added to reduce the reconstruction error. As the different distributions of different datasets influence the weights of loss terms, we set $\lambda_{s}=0.01$, $\lambda_{Irec}=0.25$, $\lambda_{a}=0.15$ and $\lambda_{adv}=0.0001$ for the DPR dataset~\cite{zhou2019deep}. }

Our training uses $0.001$ as the learning rate for training stage 1, and $0.0001$ for training stage 2 both with a batch size of 8, and the optimizer is Adam. {We conducted all the experiments on a platform with CPU i5-9400F, 16G RAM and a single NVIDIA GPU with 11GB of RAM. The results provided in our paper are generated from a model spending about 250K/30K training iterations at the first/second stage.} To better exhibit the effect of re-rendering, we utilize a Poisson blending algorithm~\cite{perez2003poisson} to combine the rebuilt faces with the backgrounds. 


\section{Experiments}

In this section, we evaluate our proposed method and compare it with state-of-the-art competitors on 5 different datasets including DPR~\cite{zhou2019deep}, FFHQ~\cite{karras2019style}, Photoface~\cite{zafeiriou2011photoface}, CelebA~\cite{liu2015deep} and Georgia Tech (GT)~\cite{gt2007data}. DPR~\cite{zhou2019deep} contains 138135 relit images that are generated by CelebA~\cite{liu2015deep} under various lighting conditions (including harsh lighting). It is used to verify our method can apply the extreme condition. FFHQ~\cite{karras2019style} consists of 70,000 high-quality images with a range of age and ethnicity.  It can be used to put our algorithm through its paces on a larger scale. Photoface~\cite{zafeiriou2011photoface} contains images of the same person under different lighting conditions, which we use to 
assess the ability in normal and albedo prediction. We also test our method on real unconstrained face images from the GT dataset~\cite{gt2007data}, which has multiple pictures of the same person with different facial expressions, lighting and poses. Furthermore, FFHQ~\cite{karras2019style} and CelebA~\cite{liu2015deep} do not have paired images, and we test the generalization capability of our method using a model trained on DPR~\cite{zhou2019deep}. In order to evaluate our method quantitatively and qualitatively, we also generate synthetic paired image data by randomly selecting of DPR lighting~\cite{zhou2019deep}, normal and albedo from SfSNet~\cite{sengupta2018sfsnet}.

\subsection{Comparison with state-of-the-art methods}

\begin{figure}[t]
	\begin{center}
		\begin{overpic}[width=\linewidth]{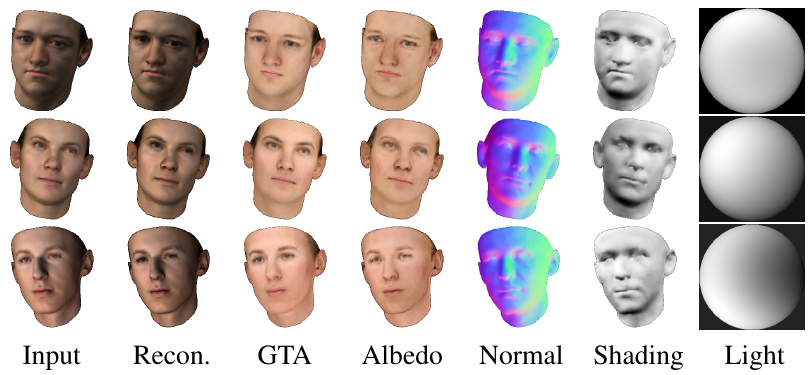}
		\end{overpic}
	\end{center}
	\vspace{-14pt}
	\caption{{{Results on synthetic data}. We compare our results with the ground truths. From left to right, they are the inputs, reconstruction faces (Recon.), ground truths of albedo maps (GTA), predicted albedo maps, predicted normal maps, shading maps and predicted light, respectively.}}
	\label{SyPairComparison}
\end{figure}

\noindent\textbf{Evaluation on synthetic data.} To quantitatively compare our model with prior works, we generate synthetic paired image data by randomly selecting lights from the DPR dataset~\cite{zhou2019deep}, normal and albedo from the SfSNet dataset~\cite{sengupta2018sfsnet}.
Table~\ref{Aligned_SyComp} reports the numerical results of our reconstruction and predicted albedo with the state-of-the-art inverse rendering method SfSNet~\cite{sengupta2018sfsnet}. The metrics used to measure the predicted albedo and reconstruction adopts MAE and RMSE. Please notice that SfSNet~\cite{sengupta2018sfsnet} is trained with the ground truth information, while ours is trained with normal initialization. As can be seen from the table, the MAE and RMSE values of our method are close to those of SfSNet~\cite{sengupta2018sfsnet} in terms of reconstruction error and albedo error. We again emphasize that our approach is able to decompose the desired components sufficiently well despite the fact that we do not train using labeled data. We provide several visual comparisons in Figure~\ref{SyPairComparison}. From the figure, it is clear to see that our results can properly decompose the normal maps while recovering reasonable albedo maps from the inputs. \\

\begin{table}[t]
	\begin{center}
		\begin{tabular}{|l|l|l|l|l|}
			\hline
			\multirow{2}{*}{Method} & \multicolumn{2}{l|}{Recon. Error} & \multicolumn{2}{l|}{Albedo Error}  \\ \cline{2-5} 
			&   MAE    &       RMSE &         MAE  &        RMSE           \\ \hline
			SfSNet      &   0.003  &      0.007 &         0.019&        0.027          \\ \hline
			Ours        &   0.014  &      0.021 &         0.051&        0.071          \\ \hline
		\end{tabular}
		\vspace{8pt}
		\caption{{Face image reconstruction and albedo reconstruction comparison between our method and SfSNet~\cite{sengupta2018sfsnet} on synthetic data in terms of MAE and RMSE.}}
		\label{Aligned_SyComp}
	\end{center}
	\begin{center}
		\scalebox{0.99}{
			\begin{tabular}{|l|c|c|c|c|}
				\hline
				Method & Mean $\pm$ std  & $<20^{\circ}$ & $<25^{\circ}$ & $<30^{\circ}$   \\
				\hline\hline
				3DMM~\cite{blanz1999morphable}         & 26.3$\pm$10.2  & 4.3$\%$   & 56.1$\%$ & 89.4$\%$ \\
				Pix2Vertex~\cite{sela2017unrestricted} & 33.9$\pm$5.6   & 24.8$\%$ & 36.1$\%$ & 47.6$\%$ \\
				SfSNet~\cite{sengupta2018sfsnet} & 25.5$\pm$9.3   & 43.6$\%$ & 57.5$\%$ & 68.7$\%$ \\
				Ours            & \textbf{23.1$\pm$8.9}  & 38.6$\%$ & 54.6$\%$ & \textbf{72.3$\%$} \\				
				\hline
				Marr Rev.~\cite{bansal2016marr}        & 28.3$\pm$10.1   & 31.8$\%$ & 36.5$\%$ & 44.4$\%$ \\
				UberNet~\cite{kokkinos2017ubernet}     & 29.1$\pm$11.5   & 30.8$\%$ & 36.5$\%$ & 55.2$\%$ \\
				NiW~\cite{trigeorgis2017face}                & 22.0$\pm$6.3     & 36.6$\%$ & 59.8$\%$ & 79.6$\%$ \\
				SfSNet-ft~\cite{sengupta2018sfsnet}       & 12.8$\pm$5.4     & {\textbf{83.7}$\%$} & 90.8$\%$ & 94.5$\%$ \\
				Ours-ft                                &\textbf{11.2$\pm$9.7}     &80.5$\%$ & \textbf{91.1}$\%$ & \textbf{96.4$\%$} \\
				\hline
		\end{tabular}}		
	\end{center}
	\caption{{{Normal comparison results on Photoface dataset~\cite{zafeiriou2011photoface}.} The data comes from SfSNet~\cite{sengupta2018sfsnet}. Lower is better for mean error, and higher is better for the percentage of correct pixels at various thresholds.}}
	\label{table_com_phdb}
\end{table}

\begin{figure}[t]
	\begin{center}
		\begin{overpic}[scale=0.65]{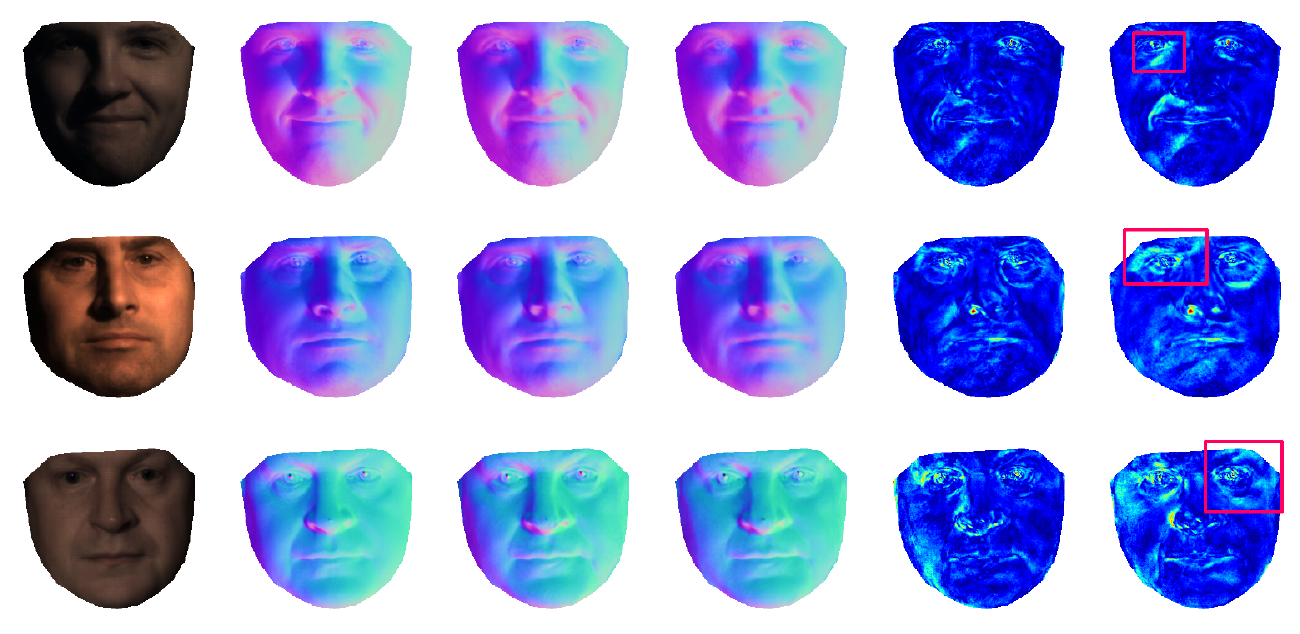}	
			\put(5,-2){\footnotesize{Input}}
			\put(22,-2){\footnotesize{GTN}}
			\put(38,-2){\footnotesize{OurN}}
			\put(53,-2){\footnotesize{SfSNetN}}
			\put(72,-2){\footnotesize{OurE}}
			\put(86,-2){\footnotesize{SfSNetE}}
		\end{overpic}
	\end{center}
	\vspace{-6pt}
	\caption{{Normal comparison with SfSNet~\cite{sengupta2018sfsnet} on the Photoface dataset~\cite{zafeiriou2011photoface}. From left to right, they are the inputs, ground truth normal maps (GTN), our normal maps (OurN), SfSNet normal maps (SfSNetN), our normal error maps (OurE) and SfSNet normal error maps (SfSNetE).}}
	\label{Fig_PhoDB}
	\begin{center}
		\begin{overpic}[scale=0.99]{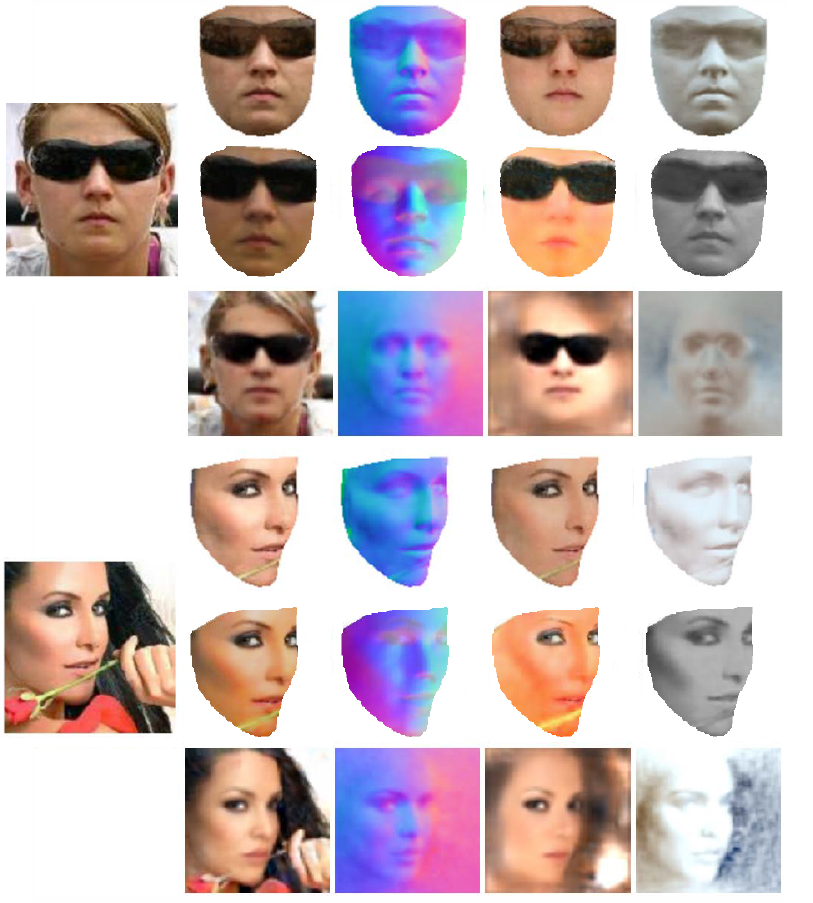}		
			\put(6,-3){\scriptsize {Input}}
			\put(24,-3){\scriptsize {Recon.}}
			\put(40,-3){\scriptsize {Normal}}
			\put(57,-3){\scriptsize {Albedo}}
			\put(74,-3){\scriptsize {Shading}}

			\put(89, 38){\footnotesize\rotatebox{90}{SfSNet}}
			\put(89, 1){\footnotesize \rotatebox{90}{Neural Face}}
			\put(89, 23){\footnotesize \rotatebox{90}{Ours}}
			
			\put(89, 88){\footnotesize\rotatebox{90}{SfSNet}}
			\put(89, 52){\footnotesize \rotatebox{90}{Neural Face}}
			\put(89, 73.5){\footnotesize \rotatebox{90}{Ours}}
		\end{overpic}
	\end{center}
	\vspace*{-5pt}
	\caption{Inverse rendering comparison with the state-of-the-art methods. Our results compare with the SfSNet~\cite{sengupta2018sfsnet} and Neural Face~\cite{shu2017neural} on the data showcased by the authors. We outperform the unsupervised method Neural Face~\cite{shu2017neural}, which regards as the baseline. It should be noted that our model does not train on the CelebA dataset~\cite{liu2015deep} due to a lack of paired images. Instead, we employ the model trained on the DPR dataset~\cite{zhou2019deep}.}
	\label{Fig_com_sfs_nf}
\end{figure}

\noindent\textbf{Evaluation of normal estimation.} We compare the quality of our estimated normals with the state-of-the-art methods, which recover from a single image. Since we do not know the split of the training dataset, we select all faces of the same person and make the permutation of images to constitute pair of images without repetition. Then we randomly split the data for our training and testing. The evaluation is the mean angular error and percentage of pixels under angular error thresholds~\cite{trigeorgis2017normal}. For a fair comparison, we also train our model on real face images. The compared model has been trained with a mixture of synthetic data and FFHQ dataset~\cite{karras2019style}. From Table~\ref{table_com_phdb}, the mean and std of predicted normal by `Ours' (not trained on the Photoface dataset~\cite{zafeiriou2011photoface}) outperform those of the others, we also show the normal error maps comparison in Figure~\ref{Fig_PhoDB}. {Figure~\ref{Fig_PhoDB} and Table~\ref{table_com_phdb} show that our predicted normal is slightly inferior to the compared methods on angle error below $20^{\circ}$, while our model clearly outperforms the competitors when the angle error is above $25^{\circ}$. This is because coarse 3DMM normal initialization may misguide our normal estimation when angles are small in the early steps.} When our model acquires a reasonable decomposition ability, we refine our model with the predicted normal instead of initialization.

{Similar to SfSNet~\cite{sengupta2018sfsnet}, we also apply the ground truth normal from Photoface dataset~\cite{zafeiriou2011photoface} to refine our model (named `Ours-ft') and use the refined model to compare with `SfSNet-ft'. More specifically, we replace the initial normal map with the ground truth normal from the Photoface, and train the model until convergence. The mean and std error of our normal is more accurate than SfSNet~\cite{sengupta2018sfsnet} from the last row in Table~\ref{table_com_phdb}. This is because our training uses paired images, which have a stronger constraint than a single. During the test, our model still retains a strong ability. The standard deviation of our result is higher than SfSNet~\cite{sengupta2018sfsnet}. We deem that our method can accurately predict the direction} {of normal in most cases without using labeled data during training. However, in some data (e.g., image pairs with a great difference between illumination), it cannot accurately estimate albedo, leading to poor shading, further influencing the final normal prediction. As a result, the outlier data make our standard deviation bigger than the labeled data trained method, SfSNet~\cite{sengupta2018sfsnet}.} \\

\begin{figure}[t]
	\begin{center}
		\centering	
		\begin{overpic}[scale=0.42]{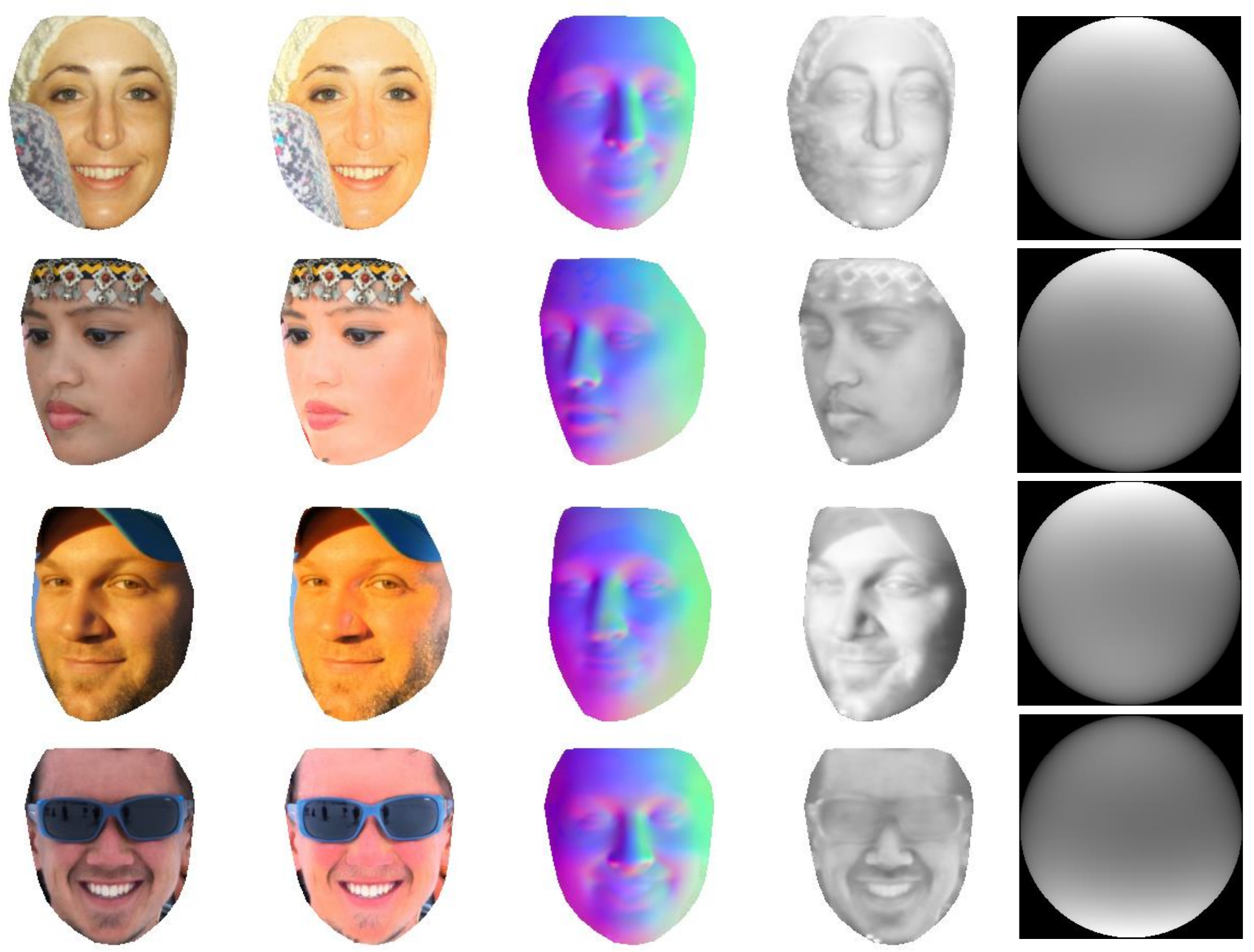}		
			\put(5, -3){\footnotesize{Input}}
			\put(24, -3){\footnotesize{Albedo}}
			\put(45, -3){\footnotesize{Normal}}
			\put(66, -3){\footnotesize{Shading}}
			\put(86, -3){\footnotesize{Light}}												
		\end{overpic}
		\caption{{Inverse rendering results on FFHQ~\cite{karras2019style} with occlusions, such as sunglasses and scarf.}}		
		\label{Fig:occlu}
	\end{center}
\end{figure}

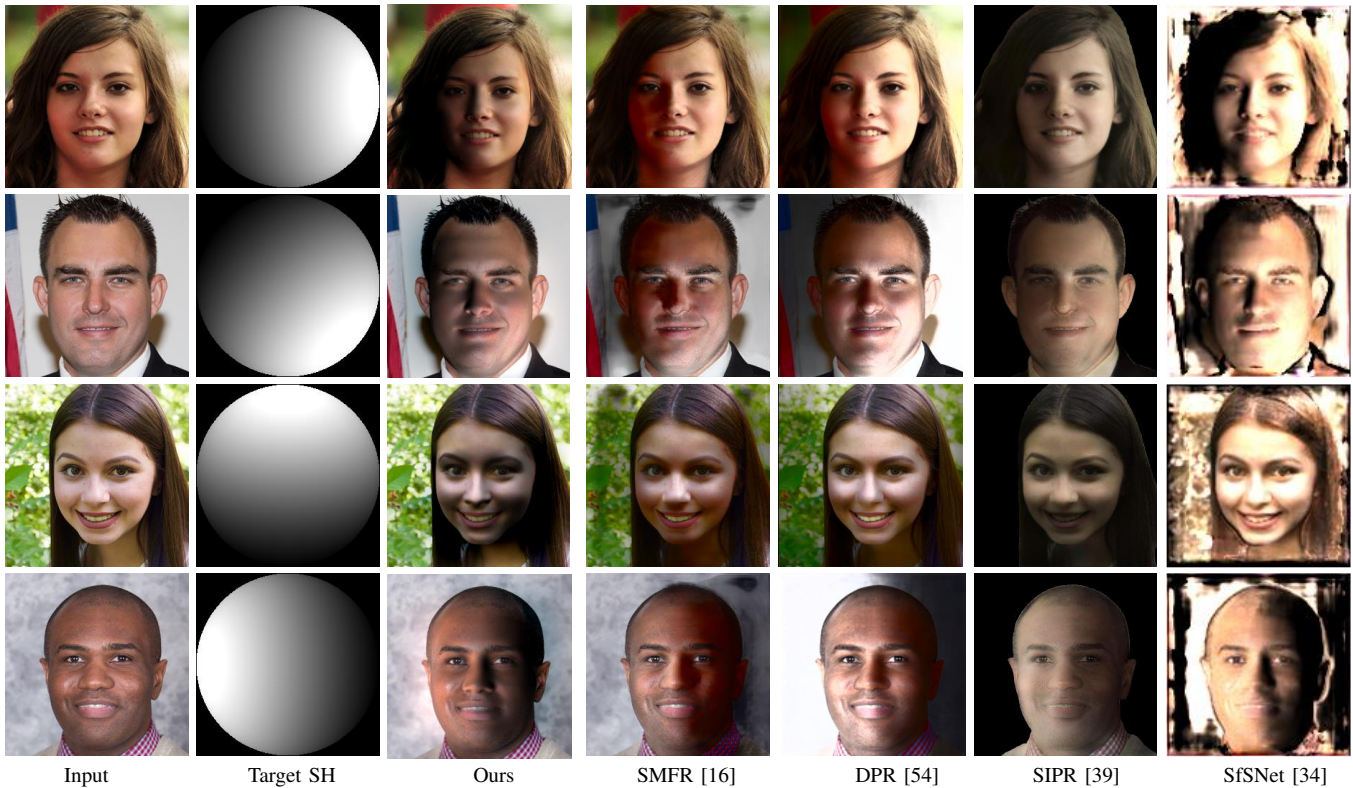
\begin{figure*}[t]
	\begin{center}
		\begin{overpic}[width=\textwidth]{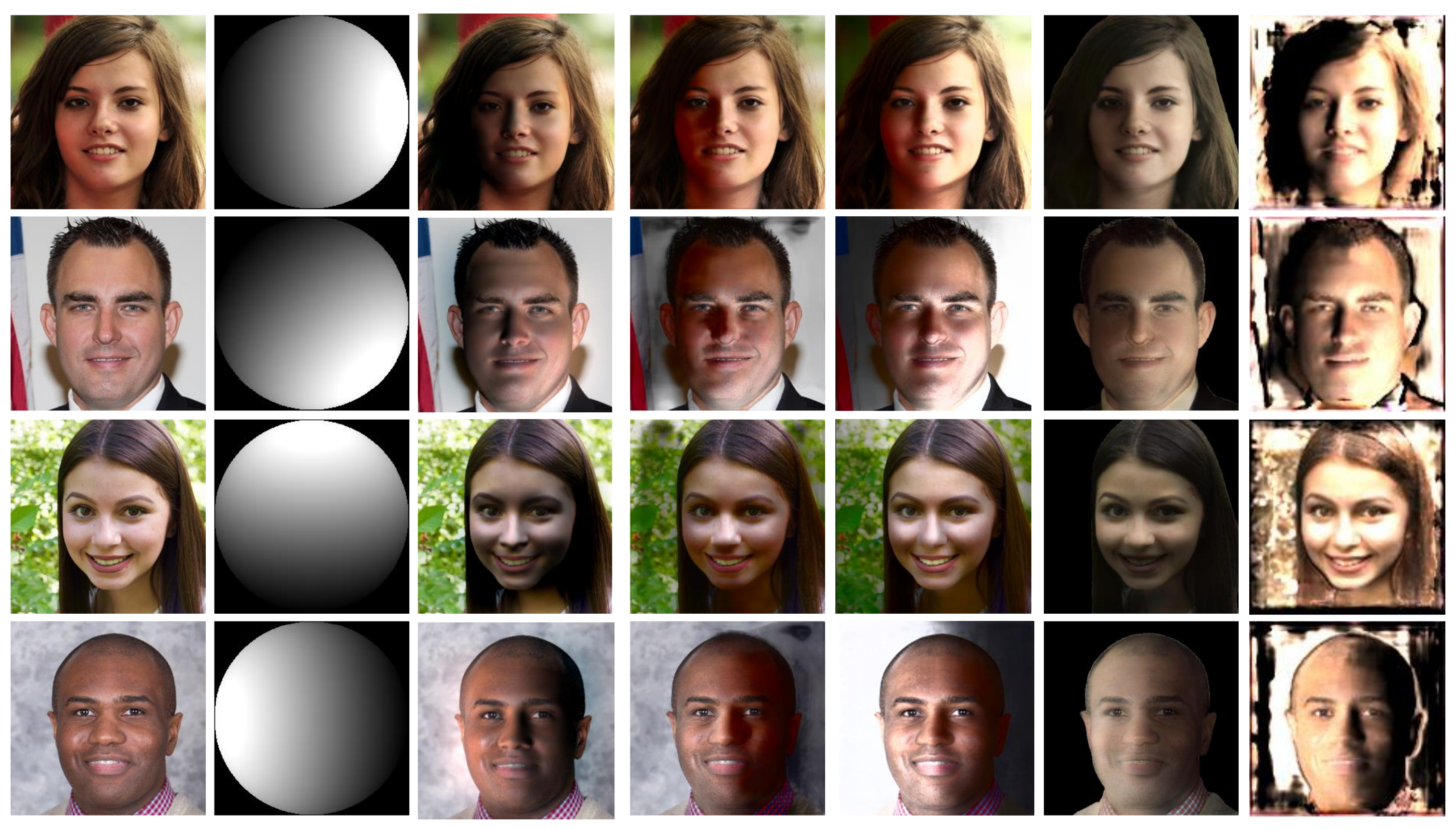}
			\put(5,-1){\footnotesize {Input}}
			\put(18.5,-1){\footnotesize {Target SH}}
			\put(35,-1){\footnotesize {Ours}}
			\put(47,-1){\footnotesize {SMFR~\cite{hou2021towards}}}
			\put(63,-1){\footnotesize {DPR~\cite{zhou2019deep}}}
			\put(76,-1){\footnotesize {SIPR~\cite{sun2019single}}}
			\put(90,-1){\footnotesize {SfSNet~\cite{sengupta2018sfsnet}}}
		\end{overpic}
		\vspace{-12pt}
		\caption{Relighting comparison with the state-of-the-art methods. Our results compare with the data provided by SMFR~\cite{hou2021towards} in qualitative on FFHQ~\cite{karras2019style}. Our model produces more natural and real cast shadows than prior works, especially around the nose, lip, and eyebrow.}
		\label{figureDPR_relight}
	\end{center}
\end{figure*}
\noindent\textbf{Evaluation of light estimation.} {For lighting evaluation, SfSNet~\cite{sengupta2018sfsnet} used 27-dimensional spherical harmonic coefficient vectors as their light. Following DPR~\cite{zhou2019deep}, the predicted lighting of our model is 9-dimensional spherical harmonic coefficient vectors. The lighting comparison with MAE/RMSE is unfair under different dimensions. For this reason, we regard lighting evaluation as a classification problem, as LDAN~\cite{zhou2018label} and SfSNet~\cite{sengupta2018sfsnet} did. Similar to SfSNet~\cite{sengupta2018sfsnet}, we evaluate the accuracy of estimated lighting with the accuracy of lighting classification. We compare the synthetic data after light clustering by K-Means.} Specifically, we use K-Means to give lights to cluster 10 classes and then compare the lighting classification correctness. Ours accuracy is 89.46\%, and SfSNet~\cite{sengupta2018sfsnet} is 94.32\%. The comparisons show that our method is slightly less effective than the supervised method, SfSNet~\cite{sengupta2018sfsnet}. It is quite usual that we have a few percentage points less than SfSNet~\cite{sengupta2018sfsnet}.\\

\noindent\textbf{Inverse rendering comparison.} In Figure~\ref{Fig_com_sfs_nf}, we show our results compared with the state-of-the-art inverse rendering methods on CelebA~\cite{liu2015deep}. It can be seen that our method can produce high-frequency albedo maps and normal maps than `Neural Face'~\cite{shu2017neural}, which is regarded as the baseline. Compared with SfSNet~\cite{sengupta2018sfsnet} trained with synthetic data and real data, our results were able to produce similar results to theirs. Significantly, the distribution gap between CelebA~\cite{liu2015deep} and DPR~\cite{zhou2019deep} leads to relatively poor reconstruction results of our method, but it shows that our model has a general ability to disentangle face images.

{In Figure~\ref{Fig:occlu}, we show that our model can be applied to a broader range of scenarios, such as a face with glasses and scarves. This is because the coarse normal is simple without occlusion, and our model learns an ability to eliminate the occlusion parts.} \\

\begin{figure*}
\begin{center}
	\begin{overpic}[scale=0.33]{imgs/SP_GT-4.pdf}
		\put(5,0){\footnotesize \color{black}{Input}}
		\put(17.5,0){\footnotesize \color{black}{Texture}}
		\put(30,0){\footnotesize \color{black}{Albedo}}
		
		\put(42.5,0){\footnotesize \color{black}{Normal}}
		\put(55.5,0){\footnotesize \color{black}{Recon.}}
		\put(66,0){\footnotesize \color{black}{Target SH}}
		
		\put(78.5,0){\footnotesize \color{black}{Shading}}
		\put(92.5,0){\footnotesize \color{black}{Relit}}
	\end{overpic}
	\vspace{-5pt}
	\caption{Decomposition and relighting results on real unconstrained images from the GT dataset (GT)~\cite{gt2007data}. Our relit faces keep local facial details with natural cast shadows under the target SH, particularly around the nose.}
	\label{GT_Decom}
	\vspace{2mm}
	\begin{overpic}[scale=0.35]{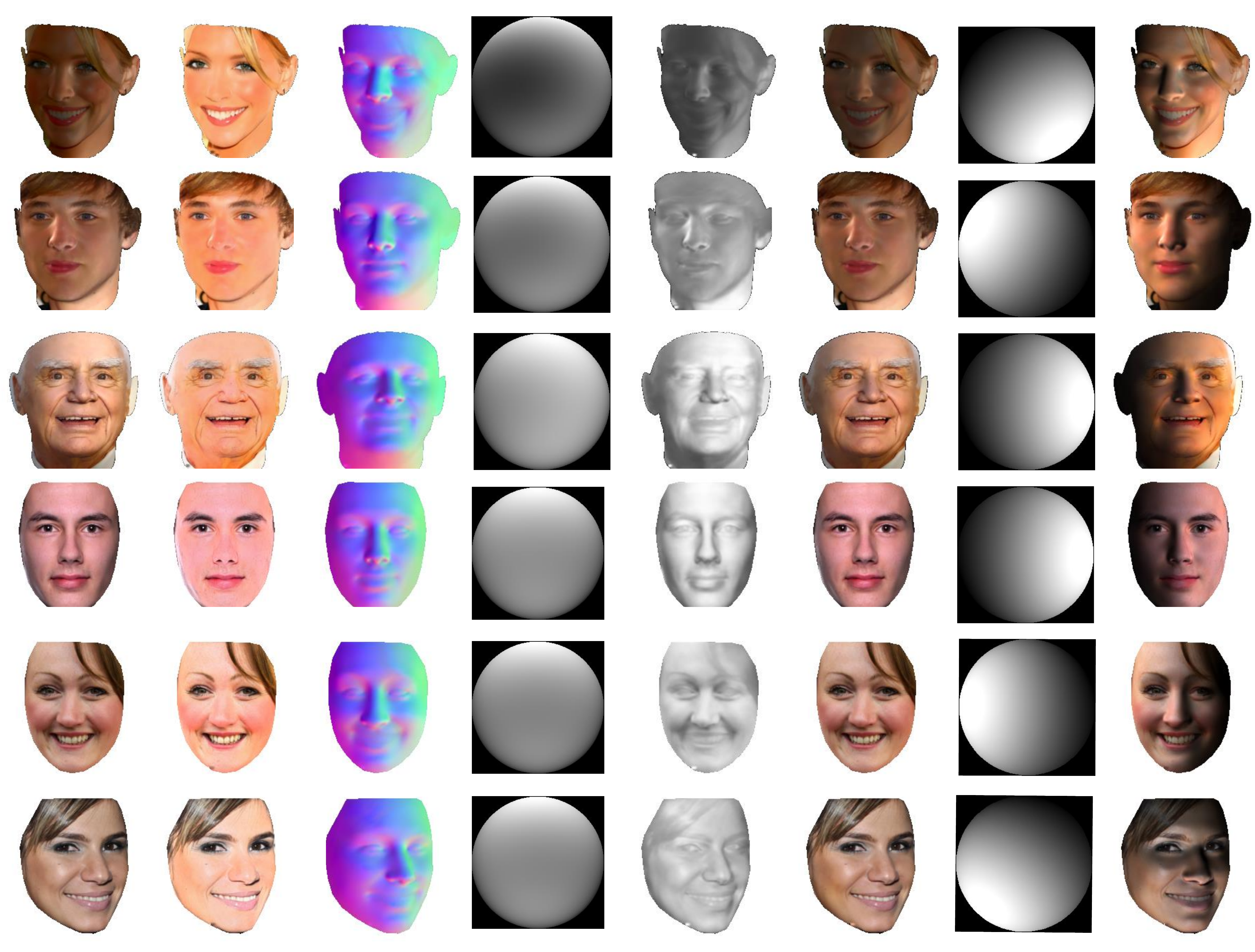}
		\put(5,-2){\footnotesize {Input}}
		\put(16,-2){\footnotesize {Albedo}}
		\put(28,-2){\footnotesize {Normal}}
		\put(41,-2){\footnotesize {Light}}
		\put(54,-2){\footnotesize {Shading}}
		\put(67,-2){\footnotesize {Recon.}}
		\put(78.5,-2){\footnotesize {Target SH}}
		\put(92.5,-2){\footnotesize {Relit}}
	\end{overpic}
	\vspace{4pt}
	\caption{Decomposition and relighting results on DPR~\cite{zhou2019deep} (first three rows) and FFHQ~\cite{karras2019style} (last three rows). Our albedo maps and normal maps with the target SH can relight new faces. The first three lines are DPR~\cite{zhou2019deep} results, and the last three lines are FFHQ~\cite{karras2019style} results.}
	\label{DPR_relight}
\end{center}
\end{figure*}
\noindent\textbf{Relighting comparison.} Figure~\ref{figureDPR_relight} shows the portrait relighting results compare with the state-of-the-art methods SMFR~\cite{hou2021towards}, DPR~\cite{zhou2019deep}, SIPR~\cite{sun2019single} and SfSNet~\cite{sengupta2018sfsnet} on FFHQ~\cite{karras2019style}. Compared to SMFR~\cite{hou2021towards}, our results produce a more realistic lighting effect on the faces because the light on face changes gradually to appear realistic without high-contrast shadows. As we all know, Single-sided glare increases the intensity of ambient light, which in turn affects the effect of light action. The high-contrast shadows on face only happen in the photography studio. Therefore, it would not be easy to see high-contrast shadows on face in the nature under the strong bright light.
Furthermore, our model is a physical-based inverse rendering model, while the shadows produced by SMFR~\cite{hou2021towards} is controlled by thresholds, which is the peak at the center and smoothly decays with the distance. An incorrect threshold will lead to unrealistic results. Give an example of shadows in the third row of the images, SMFR~\cite{hou2021towards} concerns the high-contrast shadow effects on the nose. They do, however, disregard shadow effects on the lower lip.
This erroneous lighting could be caused by the soft-shadow thresholds that are outside of the range. The outside thresholds will not work properly and will result in misjudgment results. The physical-based inverse rendering model based on lighting and geometry can accurately show the light effects on geometry. Our physical-based inverse rendering model can accurately calculate the interaction between lighting and normal. As a result, our relit faces can portray the shading more fully to provide realistic cast shadow effects. We can see from the comparison that our relighting outcomes are more natural and realistic results, as shown in Figure~\ref{figureDPR_relight}. In terms of light intensity and shadows, our illuminated faces outperform others. 

\begin{figure}[h]
	\begin{center}
		\begin{overpic}[width=\linewidth]{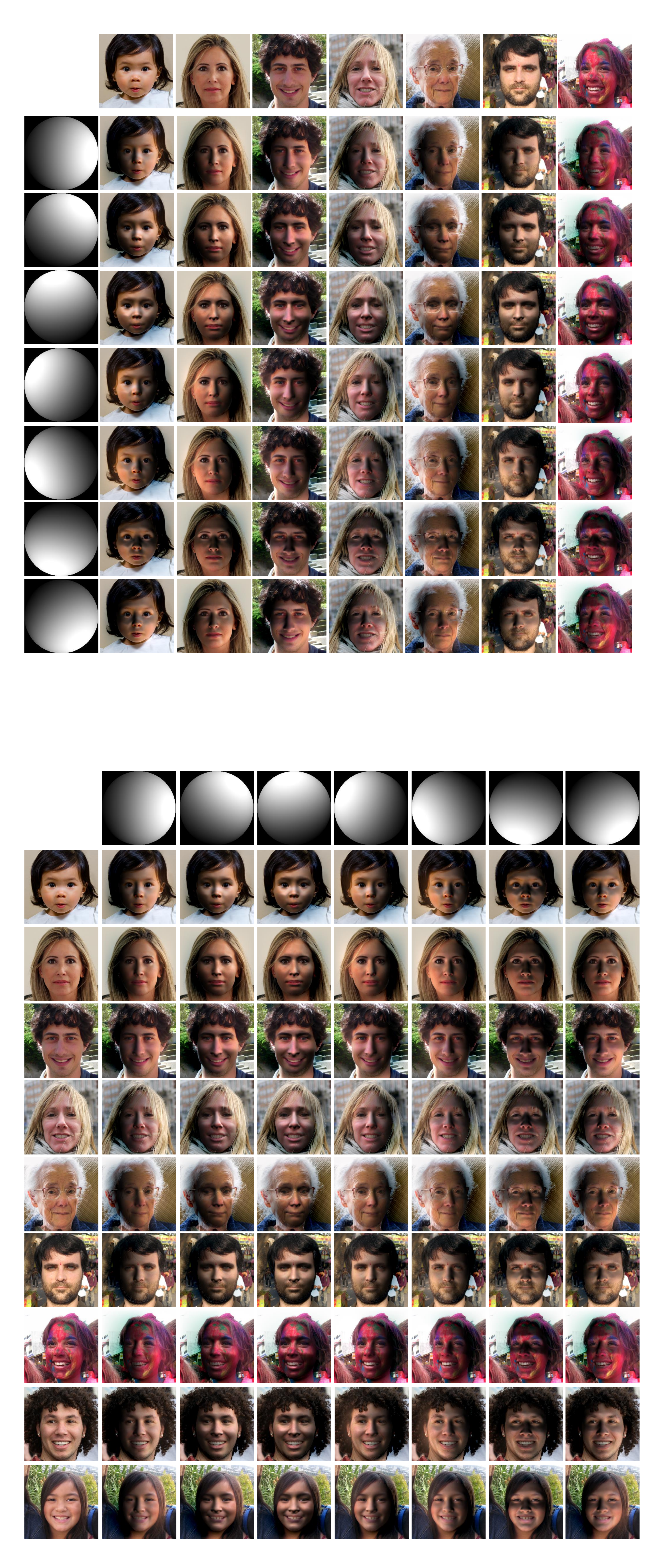}
			\put(2.5,-2){\footnotesize {Input}}
			\put(12,-2){\footnotesize {Relit 1}}
			\put(22,-2){\footnotesize {Relit 2}}
			\put(32,-2){\footnotesize {Relit 3}}
			\put(42,-2){\footnotesize {Relit 4}}
			\put(51.5,-2){\footnotesize {Relit 5}}
			\put(62,-2){\footnotesize {Relit 6}}
			\put(72,-2){\footnotesize {Relit 7}}
		\end{overpic}
		\caption{More relighting results under varied lighting conditions by our method on samples from the FFHQ~\cite{karras2019style}. The input and 7 relit faces are shown from left to right. (Best viewed by zooming in.)}
		\label{FFHQ_Relit}
	\end{center}
\end{figure}

Furthermore, we provide several results on the real face dataset, \emph{i.e.} the GT dataset~\cite{gt2007data}. Figure~\ref{GT_Decom} shows the results of our method on unconstrained real face images. The predicted albedo maps, normal maps, and lighting can reconstruct the original images and ensure that the reconstructed images obtain high-frequency details. {However, unwarping and warping processes~\cite{tran2018nonlinear} will lose the face details due to the bilinear sampling related operation. Considering that the shading should be piece-wise smooth, we assume the loss caused by warping and unwarping~\cite{tran2018nonlinear} goes to the shading map}. Thus, we use the input face $I$ and predicted shading $S$ to compute the albedo map and regard the computed albedo maps as the final output to relight a new face. More specifically, the final albedo map can be computed by $A=I/S$ under the intrinsic decomposition model. And we regard the $A$ as the final albedo maps for relighting. In Figure~\ref{GT_Decom}, It is obvious that our method can predict the albedo maps with high-frequency. On the other hand, the high-frequency relit faces under a new light with the natural cast shadows demonstrate that our method is capable of decomposing each component accurately. 

In addition, we present numerous face decomposition results on DPR~\cite{zhou2019deep} and FFHQ~\cite{karras2019style} in Figure~\ref{DPR_relight}. We also merge these decomposed components with the new light to obtain the new relit faces, as shown in Figure~\ref{DPR_relight}. We can see that our technique can properly estimate each component; \emph{e.g.}, the predicted shading maps demonstrate the correctness of our predicted light. Besides, the new relit faces under the target SH can produce natural and realistic cast shadows. The decomposition components by the effect of new light can recompose natural cast shadows, indicating that our method is proper. \\

\noindent\textbf{Runtime comparison.} {The official codes of Neural Face~\cite{shu2017neural} and SfSNet~\cite{sengupta2018sfsnet} are Lua and Matlab, respectively. A direct comparison is not fair. For this reason, we only compared with an unofficial implementation of Pytorch-based SfSNet\footnote{https://github.com/bhushan23/SfSNet-PyTorch}~\cite{sengupta2018sfsnet}. Without code optimization, the mean prediction of our model and SfSNet~\cite{sengupta2018sfsnet} are 59ms and 41ms with a batch size of 8.}

\begin{figure}[t]
\begin{center}
	\begin{overpic}[scale=0.25]{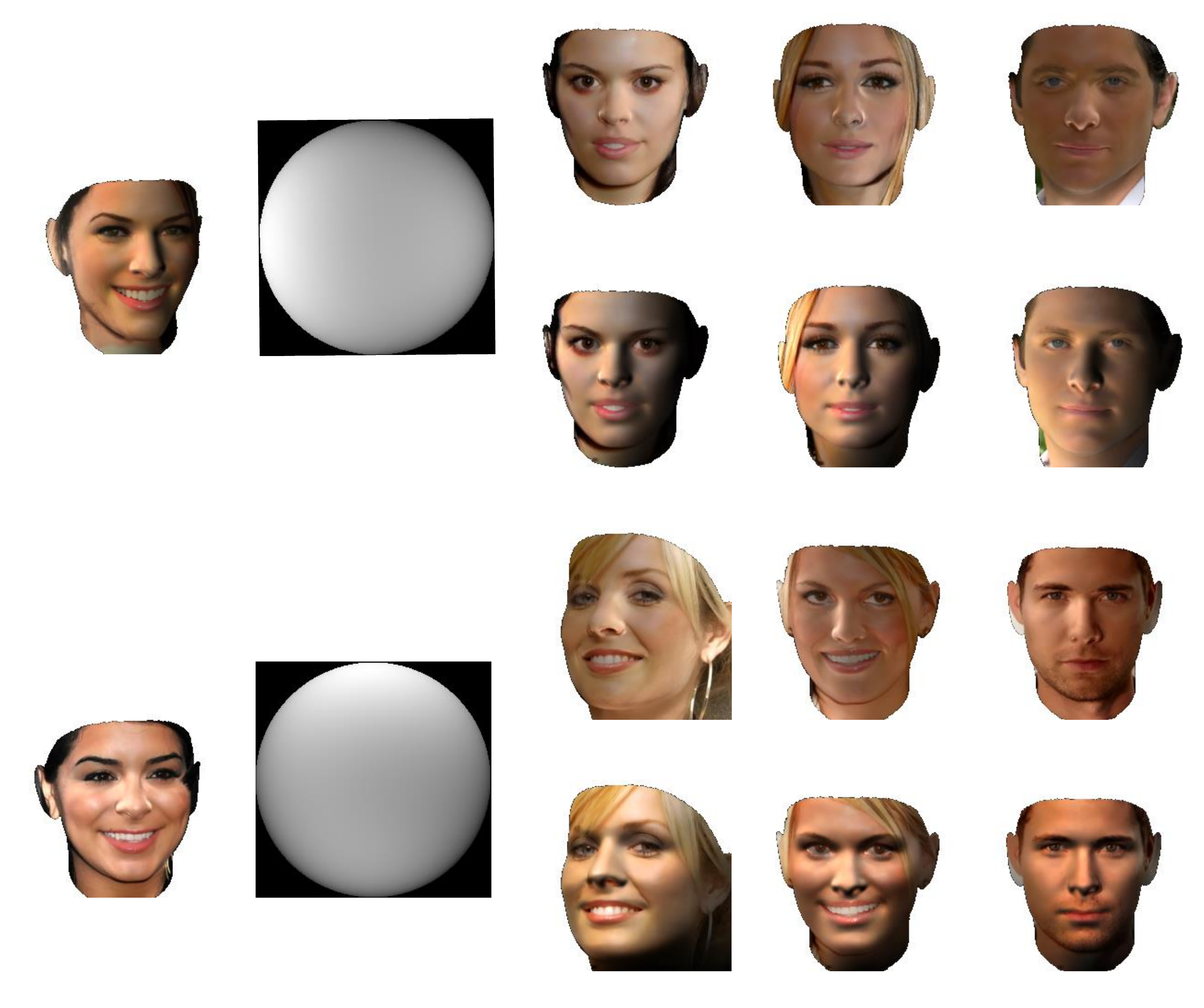}		
		\put(3,-2){\scriptsize {Source image}}
		\put(23,-2){\scriptsize {Estimated light}}
		\put(64, 20){\scriptsize {Target image}}
		\put(60,-2){\scriptsize {Transfered image}}
		\put(64,62.5){\scriptsize {Target image}}
		\put(60,41.5){\scriptsize {Transfered image}}
	\end{overpic}
\end{center}
\vspace*{-6pt}
\caption{Light transfer results from source images to the target images from the DPR dataset~\cite{zhou2019deep}.}
\label{Fig_lightTransfer}

\begin{center}
	\begin{overpic}[scale=0.4]{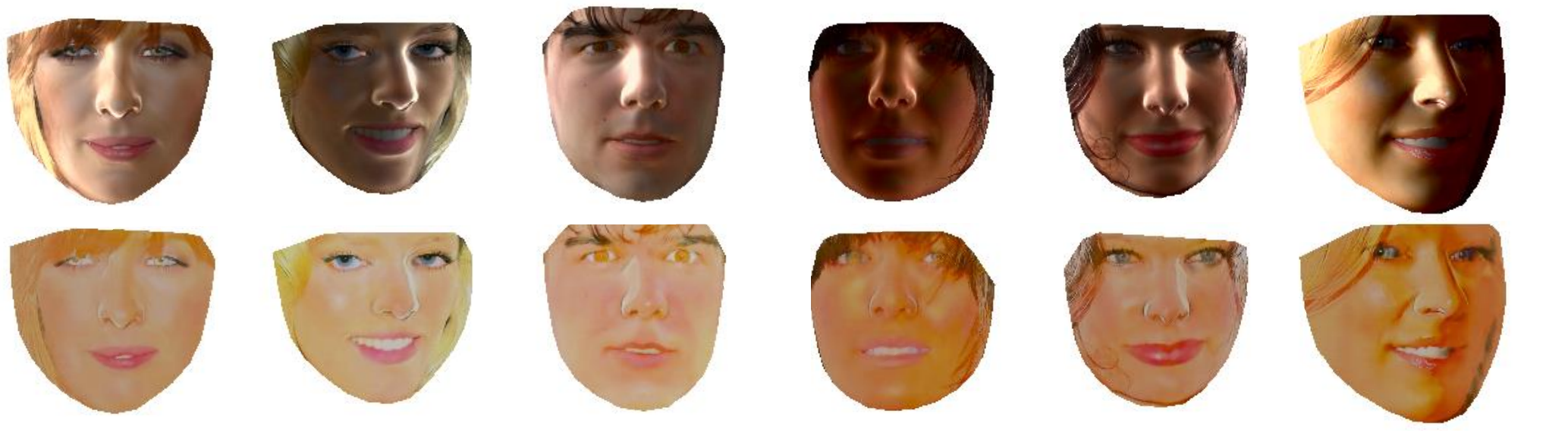}		
		\put(98,18){\scriptsize \rotatebox{90}{Input}}
		\put(98,2){\scriptsize \rotatebox{90}{De-light}}
	\end{overpic}
\end{center}
\vspace*{-10pt}
\caption{{Face de-lighting on DPR dataset~\cite{zhou2019deep}. Dlib face detector fails in face detection on the input images (Input). It can be a success after de-lighting (De-light) with our decomposition model.}}
\label{fig:delight}
\end{figure}

\begin{figure*}[t]
\begin{center}
	\begin{overpic}[width=\linewidth]{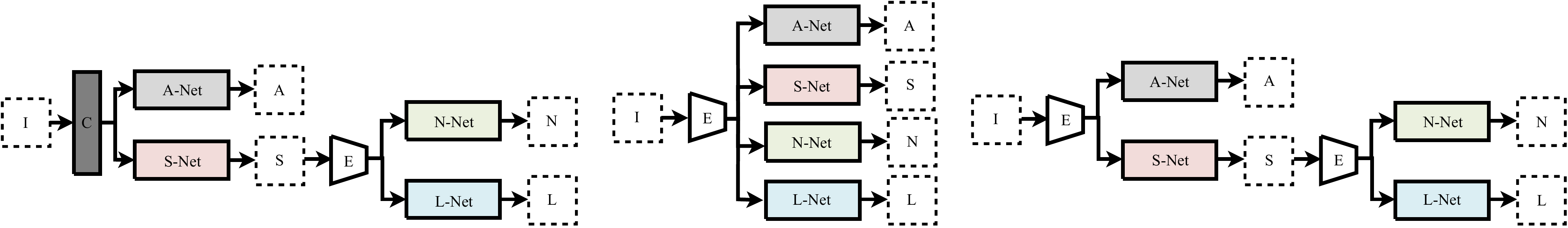}		
		\put(20,-2){\footnotesize {(a)}}
		\put(50,-2){\footnotesize {(b)}}
		\put(78,-2){\footnotesize {(c)}}
	\end{overpic}
\end{center}
\caption{{Three different decoupling network architectures. (a) Our proposed hierarchical decoupling network (\emph{HD-Net}), (b) An autoencoder network and (c) A share-encoder-based hierarchical decoupling network. The letters \emph{I, C, E, A, S, N}, and \emph{L} represent the input, convolution layer, encoder, albedo, shading, normal and lighting, respectively.}}
\label{Fig_ST}
\end{figure*}	
\subsection*{Inverse rendering applications}
In this section, we show several applications of inverse rendering. The instinctive applications are as follows:\\

\noindent\textbf{Relighting.} Figure~\ref{FFHQ_Relit}, we show a relighting application our method. It is worth mentioning that the seventh row of the face has been graffitied, our approach is still capable of properly estimating the illumination information, and the re-rendered images look natural and striking. On the other hand, it shows that our method can be applied to faces with different skin tones and accurately decompose the lighting and albedo.\\

\noindent\textbf{Light transfer.} We also exhibit light transfer results in Figure~\ref{Fig_lightTransfer}, where we apply the estimated light from the `source image' to the `target image'. The natural and realistic transfer results illustrate the accuracy of our decomposition.\\

\noindent\textbf{De-lighting.} {The accuracy of face detection is sensitive to the light on face images. We believe that another application is de-lighting (as shown in Figure~\ref{fig:delight}) to improve the accuracy of face detection. Thus, we randomly selected 60,000 face images from the extreme lighting face dataset, DPR~\cite{zhou2019deep}, to evaluate the actual application of our model. The accuracy of face detection with dlib face detector\footnote{http://dlib.net/} is $94.59\%$. After that, we utilized the dlib face detector to detect face albedo maps generated from our model, and the accuracy increased to $97.85\%$.}

\section{Ablation study} 
\begin{table}[t]
\begin{center}
	\scalebox{0.99}{
		\begin{tabular}{|c|c|c|c|c|c|c|}
			\hline
			Setting & Mean $\pm$ std  & $<20^{\circ}$ & $<25^{\circ}$ & $<30^{\circ}$  & MAE & RMSE \\
			\hline\hline
			(a)         & 8.9 $\pm$11.6   & 87.4$\%$   & 93.5$\%$    & 96.5$\%$   & 0.051  & 0.071\\
			(b)         & 9.0$\pm$12.0    & 86.4$\%$   & 93.1$\%$    & 96.3$\%$   & 0.053  & 0.074\\
			(c)         & 10.1$\pm$13.4   & 81.9$\%$   & 90.0$\%$    & 94.3$\%$   & 0.056  & 0.083\\
			\hline
	\end{tabular}}	
	\vspace*{3pt}	
	\caption{{Normal and albedo (MAE/RMSE) evaluation on Photoface dataset~\cite{zafeiriou2011photoface} of different network architectures.}}
	\label{table:Table_ablationNet}
\end{center}
\end{table}

In the following, we perform several ablation studies to explore different aspects of our approach in more detail.\\

\noindent\textbf{Ablation of network architectures.} {To demonstrate the effectiveness of our hierarchical decoupling network, we train two additional models based on Lehtinen~\emph{et al.}~\cite{lehtinen2018noise2noise} (see Figure~\ref{Fig_ST}). The results are shown in Table~\ref{table:Table_ablationNet}. From the table, it can be seen that the accuracy of normal and individual albedo components predicted by \emph{HD-Net} (a) is much better than the other two ((b) and (c)). The comparisons confirm the effectiveness of our hierarchical design. In addition, the number of trainable parameters is used to evaluate the complexity of a model. For neural networks, the higher the number of parameters, the more complex the model. Here, we report the complexity of our model with the number of trainable parameters of our model. The models of (b) and (c) have approximately 6.8M parameters and 7M parameters, whereas our \emph{HD-Net} model has about 5M parameters, which is pretty small. We can achieve better results with a smaller model than the other two network architectures. When the batch size is 8, our \emph{HD-Net} only requires 6.3GB of GPU memory on $3\times256\times256$ inputs. }

\begin{figure}[t]
\begin{center}
	\begin{overpic}[scale=0.9]{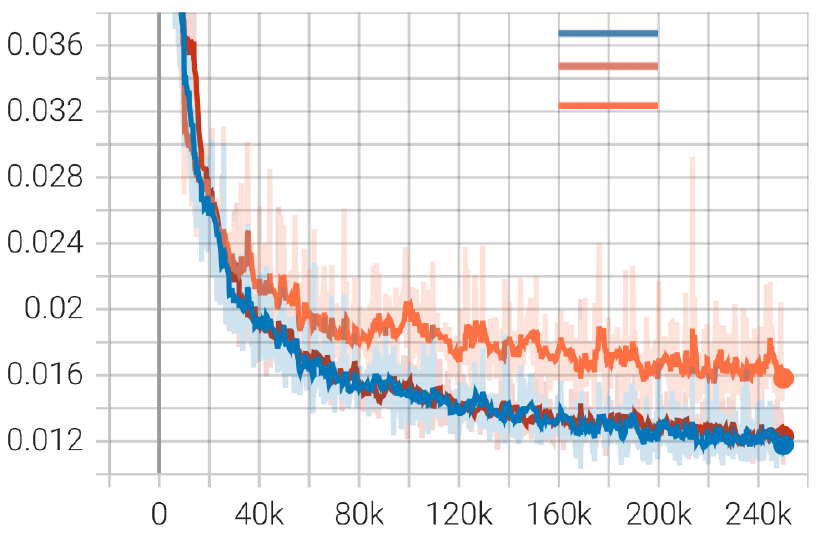}	
		\put(82,60.5){\footnotesize{w/o D}}
		\put(82,56){\footnotesize{w/ DL}}
		\put(82,51){\footnotesize{w/ DLS}}
	\end{overpic}
	\vspace*{-6pt}
	\caption{{The normal loss of different detaching settings. {w/o D, w / DLS and w / DL stand for training networks without predicted shading and computed light detaching, with predicted shading and computed light detaching, and with computed light detaching, respectively.}}}
	\label{Fig:NLoss}
\end{center}
\end{figure}

To ease the one-to-three decomposition task, we employ a similar network architecture from~\cite{zhang2019kindling} for our shading prediction. The benefits of splitting shading prediction network \emph{S-Net} are twofold: one is to speed up the convergence of the network and reduce the training time, and the other is that the predicted shading guides our normal and lighting estimation. \\

\begin{table}[t]
\begin{center}
	\scalebox{0.99}{
		\begin{tabular}{|c|c|c|c|c|c|c|}
			\hline
			Setting & Mean $\pm$ std  & $<20^{\circ}$ & $<25^{\circ}$ & $<30^{\circ}$  & MAE & RMSE \\
			\hline\hline
			w/o D     & 9.1$\pm$11.8    & 86.8$\%$   & 93.1$\%$    & 96.1$\%$   & 0.053  & 0.079\\
			w/ DLS     & 10.7$\pm$16.1   & 73.0$\%$   & 84.9$\%$    & 91.5$\%$   & 0.046  & 0.068\\
			w/ DL      & 8.9$\pm$11.6    & 87.4$\%$   & 93.5$\%$    & 96.5$\%$   & 0.051  & 0.071\\
			\hline
	\end{tabular}}		
\end{center}
\caption{{Normal and albedo (MAE/RMSE) evaluation on Photoface~\cite{zafeiriou2011photoface} with/without detach. {w/o D, w/ DLS and w/ DL stand for training networks without predicted shading and computed light detaching, with predicted shading, and computed light detaching and with computed light detaching, respectively.}}}
\label{table:DetachRelated}
\begin{center}
	\scalebox{0.99}{
		\begin{tabular}{|c|c|c|c|c|c|c|}
			\hline
			Setting & Mean $\pm$ std  & $<20^{\circ}$ & $<25^{\circ}$ & $<30^{\circ}$  & MAE & RMSE \\
			\hline\hline
			(a)         & 8.9 $\pm$11.6   & 87.4$\%$   & 93.5$\%$    & 96.5$\%$   & 0.051  & 0.071\\
			(b)         & 9.2$\pm$12.1    & 87.3$\%$   & 93.1$\%$    & 95.6$\%$   & 0.057  & 0.081\\
			\hline
	\end{tabular}}	
	\vspace*{2pt}	
	\caption{{Normal and albedo (MAE/RMSE) evaluation on Photoface~\cite{zafeiriou2011photoface} of different training strategies. (a) and (b) are training with two steps and with an end-to-end way, respectively.}}
	\label{table:trainstrategy}
\end{center}
\end{table}

\noindent\textbf{Ablation of network detaching.} {Table~\ref{table:DetachRelated} shows the normal reconstruction error about different settings of network detaching. As can be seen from the table, detaching light (w/ DL) is able to predict a more accurate normal map, although the MAE and RMSE of albedo increase. We consider that the optimal model is the predicted normal with a smaller error. For there is no real albedo for constraint, we assume that the albedo has converged as long as the two albedo maps are consistent. In addition, it can be seen that detaching light is able to reach the convergence state faster than the other two settings in Figure~\ref{Fig:NLoss}. Obviously, detaching light is the best setting for \emph{HD-Net}.}\\

\noindent\textbf{Ablation of training strategies.} {Table~\ref{table:trainstrategy} shows the normal reconstruction and albedo error about different training strategies, (a) and (b). In addition, we also evaluate lighting classification correctness with different training strategies. The lighting classification correctness of training in two steps (a) and training with an end-to-end way (b) are $89.46\%$ and $87.23\%$, respectively.}\\

\begin{table}[t]
\begin{center}
	\scalebox{0.99}{
		\begin{tabular}{|c|c|c|c|c|c|c|}
			\hline
			Setting & Mean $\pm$ std  & $<20^{\circ}$ & $<25^{\circ}$ & $<30^{\circ}$  & MAE & RMSE \\
			\hline\hline
			w/ NC        & 10.2$\pm$13.2   & 82.6$\%$   & 90.2$\%$    & 94.4$\%$   & 0.052  & 0.077\\
			w/o AD       & 9.5$\pm$12.1    & 86.2$\%$   & 92.4$\%$    & 95.6$\%$   & 0.145  & 0.183\\
			Ours       & 8.9 $\pm$11.6   & 87.4$\%$   & 93.5$\%$    & 96.5$\%$   & 0.051  & 0.071\\
			\hline
	\end{tabular}}	
	\vspace*{3pt}	
	\caption{{Normal comparison on Photoface dataset~\cite{zafeiriou2011photoface} with different loss settings.}}
	\label{table:LossFunction}
\end{center}%
\begin{center}
	\begin{tabular}{|l|l|l|l|l|l|l|l|l|}
		\hline
		Setting & $\lambda_{Irec}$  & $\lambda_{s}$ & $\lambda_{Srec}$ & $\lambda_{a}$ & $\lambda_{n}$ & $\lambda_{l}$ & $\lambda_{adv}$   \\ 				\hline\hline
		1    & 0.25           & 0.1 & 0.01          & 0.25         & 0.25          & 0.01 & 0.001         \\ \hline
		2    & 0.25           & 0.1 & 0.01          & \textbf{0.1} & \textbf{0.5}  & 0.01 & 0.001            \\ \hline
		3    & 0.25           & 0.1 & \textbf{0.1}  & 0.25         &\textbf{0.5}   & 0.01 & 0.001          \\ \hline
		4    & \textbf{0.5}   & 0.1 & 0.01          & 0.25         & \textbf{0.5}  & 0.01 & 0.001          \\ \hline
		5    & 0.25           & 0.1 & 0.01          & 0.25         & \textbf{0.5}  & 0.01 & 0.001          \\ \hline
	\end{tabular}
	\vspace*{4pt}
	\caption{{Different weight settings of loss items.}}
	\label{table:Parameters}
\end{center}
\begin{center}
	\scalebox{0.99}{
		\begin{tabular}{|c|c|c|c|c|c|c|}
			\hline
			Setting & Mean $\pm$ std  & $<20^{\circ}$ & $<25^{\circ}$ & $<30^{\circ}$  & MAE & RMSE \\
			\hline\hline
			1     & 13.9$\pm$22.8   & 50.2$\%$   & 65.7$\%$    & 77.0$\%$   & 0.079  & 0.119\\
			2     & 12.0$\pm$17.1   & 83.3$\%$   & 90.4$\%$    & 94.3$\%$   & 0.109  & 0.149\\
			3     & 10.8$\pm$14.8   & 76.9$\%$   & 86.2$\%$    & 92.1$\%$   & 0.088  & 0.117\\
			4     & 10.2$\pm$12.9   & 83.9$\%$   & 90.3$\%$    & 94.6$\%$   & 0.082  & 0.107\\
			5     & 8.9 $\pm$11.6   & 87.4$\%$   & 93.5$\%$    & 96.5$\%$   & 0.051  & 0.071\\
			\hline
	\end{tabular}}	
	\vspace*{2pt}	
	\caption{{Normal and albedo (MAE/RMSE) evaluation on Photoface dataset~\cite{zafeiriou2011photoface} of different weight settings.}}
	\label{table:ParaSettingsEva}
\end{center}
\begin{center}
	\scalebox{0.99}{
		\begin{tabular}{|c|c|c|c|c|c|c|}
			\hline
			Noise & Mean $\pm$ std  & $<20^{\circ}y$ & $<25^{\circ}$ & $<30^{\circ}$  & MAE & RMSE \\
			\hline\hline
			GN         & 9.9$\pm$12.8    & 86.3$\%$   & 91.6$\%$    & 95.3$\%$   & 0.076  & 0.104\\
			SN         & 12.4$\pm$18.9   & 61.8$\%$   & 76.3$\%$    & 85.9$\%$   & 0.081  & 0.116\\
			Ours       & 8.9$\pm$11.6    & 87.4$\%$   & 93.5$\%$    & 96.5$\%$   & 0.051  & 0.071\\
			\hline
	\end{tabular}}
\end{center}
\caption{{Normal and albedo (MAE/RMSE) evaluation on Photoface dataset~\cite{zafeiriou2011photoface} with Gaussian and salt-pepper noises.}}
\label{table:NormalNoise}	
\end{table}

\noindent\textbf{Ablation of loss functions.} {To demonstrate the effectiveness of our loss design, we train an additional model without adversarial loss $\mathcal{L}_{adv}$ (w/o AD) and with a normal consistent loss (w/ NC) between $N_{i}$ and $N_{j}$. {As can be seen from Table~\ref{table:LossFunction}, our proposed model is able to produce significantly better results for albedo and normal prediction than the models trained without $\mathcal{L}_{adv}$ and with a normal consistent loss}. Our model outperforms the model without $\mathcal{L}_{adv}$ and with normal consistent loss. {The normal consistent loss can be regarded as a prior during training. However, it would increase the complexity of calibrating weights for balancing the influence among loss functionss.} As a trade-off, we forego incorporating normal consistent loss.}\\

\begin{figure}[t]
\begin{center}
	\begin{overpic}[width=\linewidth]{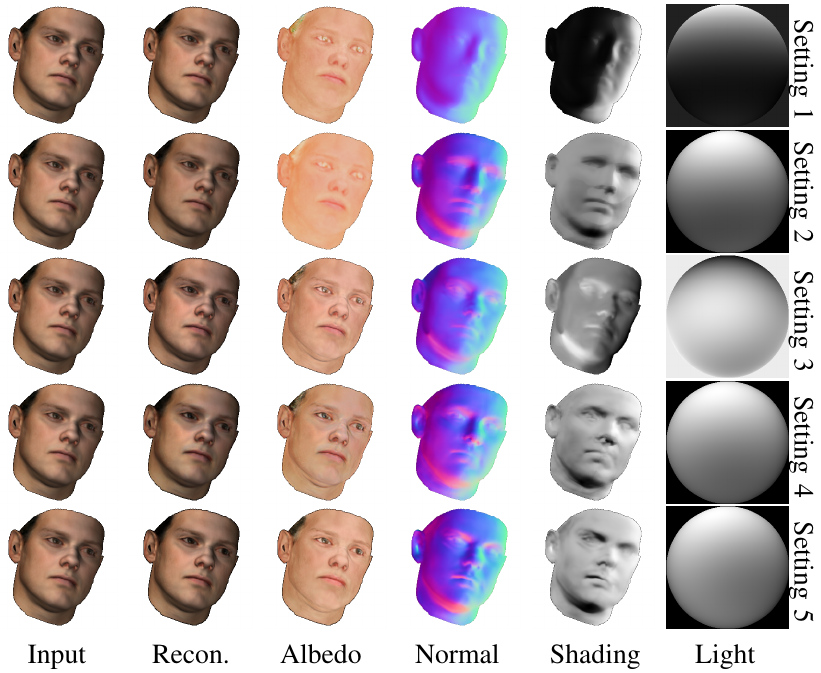}
	\end{overpic}
\end{center}
\vspace{-14pt}
\caption{{Visualized results with different weight settings of loss items.}}
\label{fig:ParaSettings}
\begin{center}
	\begin{overpic}[width=\linewidth]{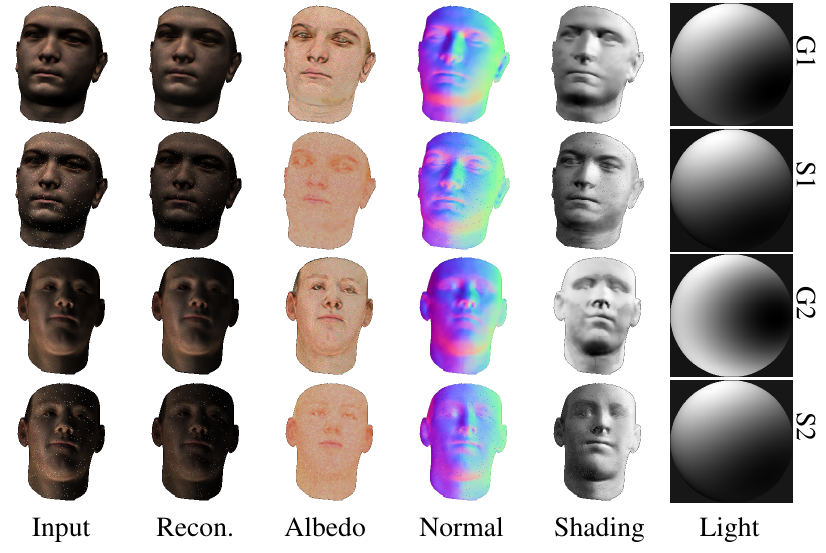}
	\end{overpic}
\end{center}
\vspace{-15pt}
\caption{{Visualized results with gaussian noises (G1/G2) and salt-peppers noises (S1/S2).}}
\label{fig:Noise}
\end{figure}

\noindent\textbf{Ablation of loss parameter settings.} {There are seven loss items to control the performance of \emph{HD-Net}, as can be seen in Table \ref{table:Parameters}, Table \ref{table:ParaSettingsEva} and Figure \ref{fig:ParaSettings}. In Figure \ref{fig:ParaSettings}, for example, Setting 1 is able to decompose the individual components, but the normal estimation deviates greatly. The predicted albedo losses some details in Setting 2, the predicted normal contains some light in Setting 3 and the reconstruction face loses the original appearance in Setting 4. With suitable weight parameters, our network is able to accomplish the task, which can be found in Setting 5.}\\

\noindent\textbf{Ablation of Gaussian/salt-pepper noises.} {We show the performance of our model trained on the face with Gaussian noises ($\sigma=0.1$) or salt-pepper noises ($SNR=0.01$) to evaluate the robustness of our method. Table~\ref{table:NormalNoise} shows the normal reconstruction and abledo error with different noise on face images, and Figure~\ref{fig:Noise} presents several visualized results of our model. Face images with Gaussian noises (G1/G2 in Figure~\ref{fig:Noise}) will treat the noise as part of the albedo, while face images with salt-pepper noises (S1/S2 in Figure~\ref{fig:Noise}) will treat the noise as part of the face geometry. Regardless of how the model handles the noise, the model is able to decompose each component. It is worth mentioning that face with noise increases training time. The model is not optimal at the current number of 250k iterations, especially with the influence of salt-pepper noises.}


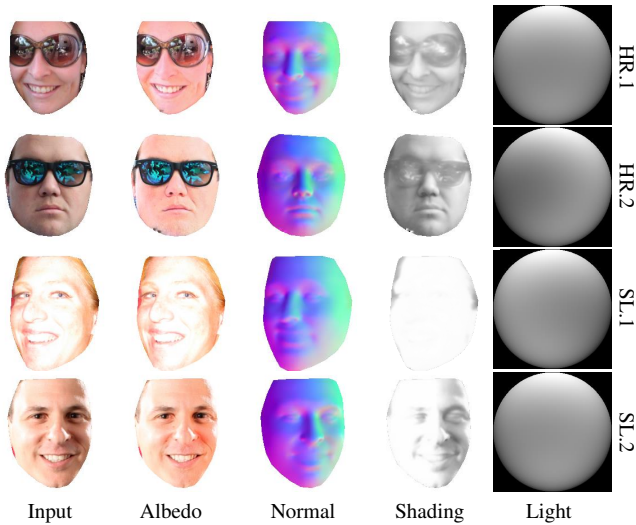
\begin{figure}[t]
\begin{center}
	\begin{overpic}[scale=0.42]{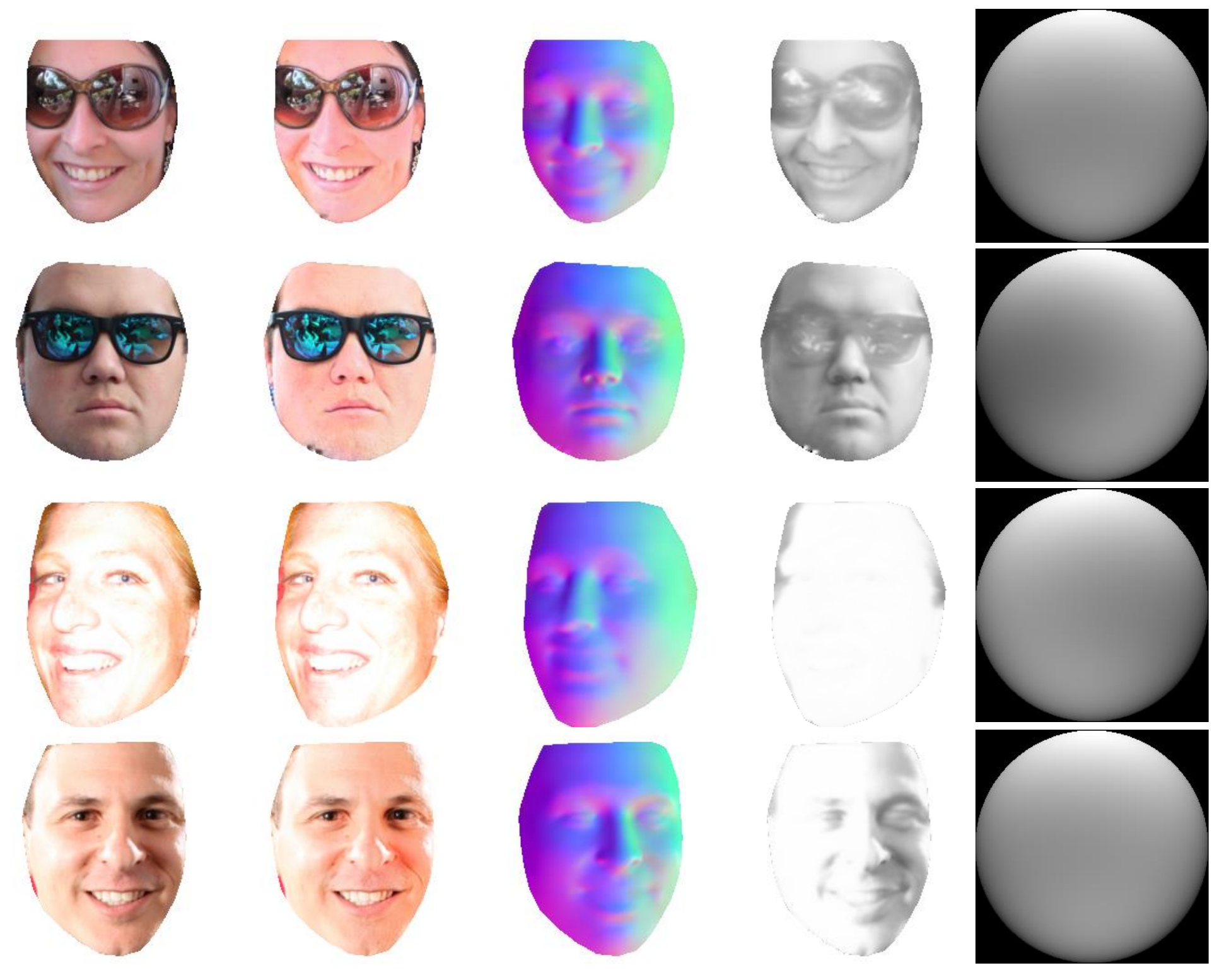}		
		\put(100, 73){\footnotesize\rotatebox{270}{HR.1}}
		\put(100, 54){\footnotesize\rotatebox{270}{HR.2} }
		\put(100, 34){\footnotesize\rotatebox{270}{SL.1} }
		\put(100, 14){\footnotesize\rotatebox{270}{SL.2} }	
		\put(5, -3){\footnotesize{Input}}			
		\put(23, -3){\footnotesize{Albedo}}			
		\put(44, -3){\footnotesize{Normal}}			
		\put(64, -3){\footnotesize{Shading}}			
		\put(85, -3){\footnotesize{Light}}			
	\end{overpic}
	\caption{{{Failure cases.} Highly reflective glasses (HR.1/HR.2), severe lighting (SL.1/SL.2).}	}
	\label{failurecase}
\end{center}
\end{figure}

\section{Concluding Remarks}
In this paper, we have presented a novel hierarchical decoupling network to solve the problem of face inverse rendering in the wild. The relief of data preparation can significantly broaden the applicable range of face inverse rendering. Our hierarchical decoupling network is inspired by a divide-and-conquer strategy, which converts a one-to-three decomposition problem into two decoupled one-to-two sub-problems. Our model can learn the implicit relationship of the light between the paired images during training, which improves its generalization ability to unseen data. The hierarchical decoupling network has revealed its advantages over other state-of-the-art inverse rendering approaches through extensive experiments. 

{The proposed method still has limitations, some of which are shown in Figure~\ref{failurecase}. These belong to extreme situations, such as a face with highly reflective glasses (HR.1/HR.2 in Figure \ref{failurecase}) or extreme lighting (SL.1/SL.2 in Figure \ref{failurecase}).

Our method can predict plausible components without ground truths. The relit faces are pretty well in different lighting directions (as shown in Figure~\ref{figureDPR_relight}, Figure~\ref{GT_Decom}, Figure~\ref{DPR_relight} and Figure~\ref{FFHQ_Relit}), while the results has some artifacts compared to original images. There are two reasons, the first one is that our model is trained with the masks. In order to compare with other methods, we utilize a Poisson blending~\cite{perez2003poisson} to rebuild relit faces with backgrounds, resulting in the inconsistency of the whole image. Second, predicted smooth shading losses the details, and it is hard for \emph{NLD-Net} to produce normal details.

{
In addition, it is worth mentioning that the unwarping and warping functions limit the quality of final results. For example, the predicted albedo is used to relight a new face with predicted normal and new light. However, warping and unwarping processes will cause detail loss because of using bilinear sampling on image pixels, thus leading to unsatisfactory quality when the warped components are applied to relight new faces.} For real face images in the wild, our network is limited by the size of warped and unwarped images, and if the size of unwarped face images is $512*512$ or larger, then the estimated face components will perform more plausible. Besides, our network does not train in an end-to-end way. Therefore, we expect future work to explore further robust face features that can be used in extreme situations. We would like to build an end-to-end model for learning 3DMM to align unwarped face pairs of images and simultaneously predict inverse renderings components.}


\ifCLASSOPTIONcaptionsoff
\newpage
\fi



%

\bibliographystyle{ieee}
\bibliography{ref}



%
\vspace{-12.5pt}
\begin{IEEEbiography}[{\includegraphics[width=1in,height=1.25in,clip,keepaspectratio]{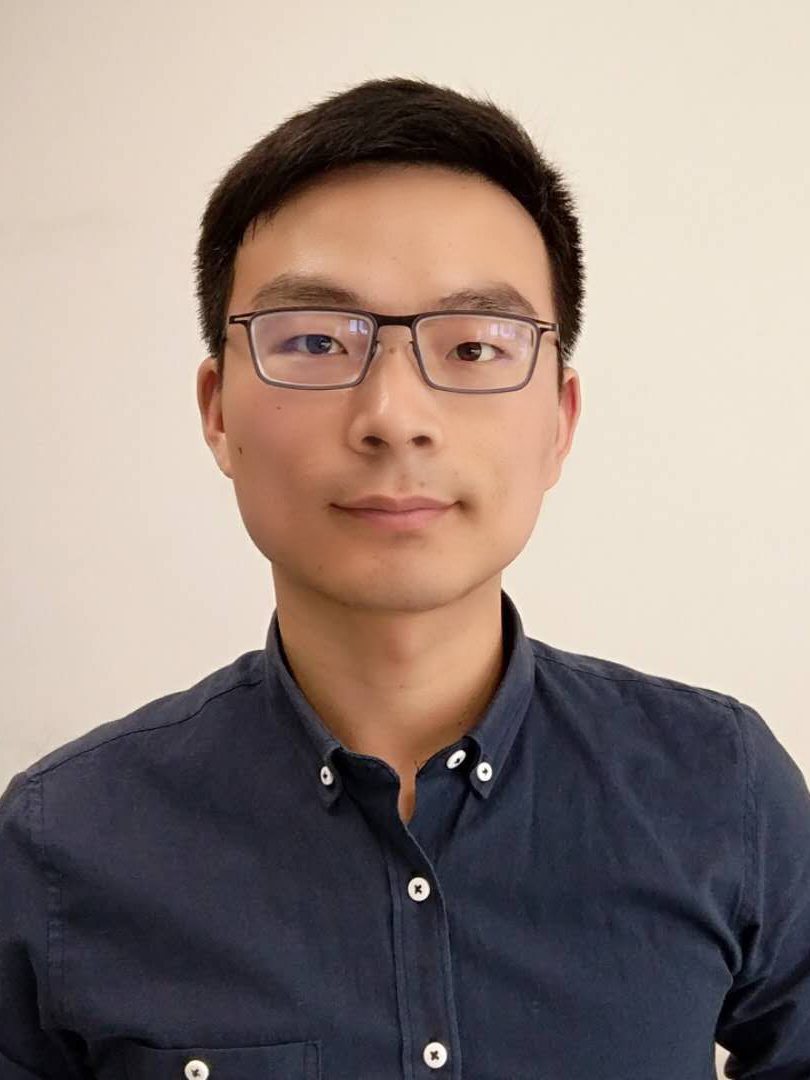}}]{Meng Wang}
	received the B.E. degree from the School of Information Science and Engineering, Henan University of Technology, Henan, China, in 2012, and the M.S. degrees in software engineering from the College of Information, Liaoning University, Liaoning, China, in 2015. He is currently pursuing the Ph.D. degree with the College of Intelligence and Computing, Tianjin University. His research interests include computer vision, machine learning, and pattern recognition.
\end{IEEEbiography}
\vspace{-12.5pt}
\begin{IEEEbiography}[{\includegraphics[width=1in,height=1.25in,clip,keepaspectratio]{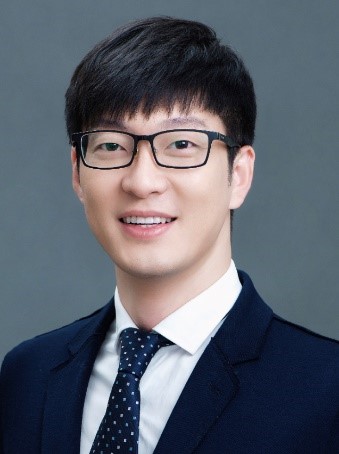}}]{Xiaojie Guo}
	(Senior Member, IEEE) is a tenured Associate Professor with the College of Intelligence and Computing, Tianjin University, Tianjin, China. Dr. Guo was a recipient of the Piero Zamperoni Best Student Paper Award in the International Conference on Pattern Recognition in 2010, the IEEE ICME Best Student Paper Runner-Up Award in 2018, the PRCV Best Student Paper Runner-Up Award in 2020.  
\end{IEEEbiography}
\vspace{-12.5pt}
\begin{IEEEbiography}[{\includegraphics[width=1in,height=1.25in,clip,keepaspectratio]{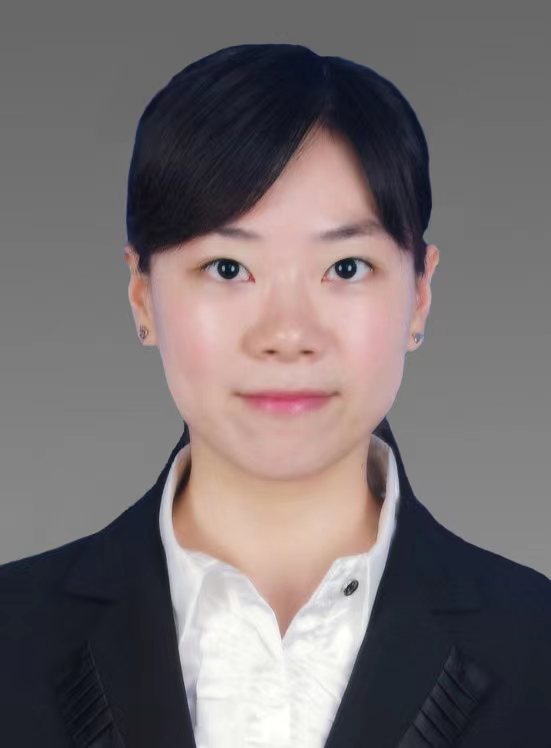}}]{Wenjing Dai}
	received the B.S. and M.S. degrees from the College of Intelligence and Computing, Tianjin University, Tianjn, China, in 2016 and 2019, respectively. Her research interests include Visualization, Visual Analytics and Machine Learning.
\end{IEEEbiography}
\vspace{-12.5pt}
\begin{IEEEbiography}[{\includegraphics[width=1in,height=1.25in,clip,keepaspectratio]{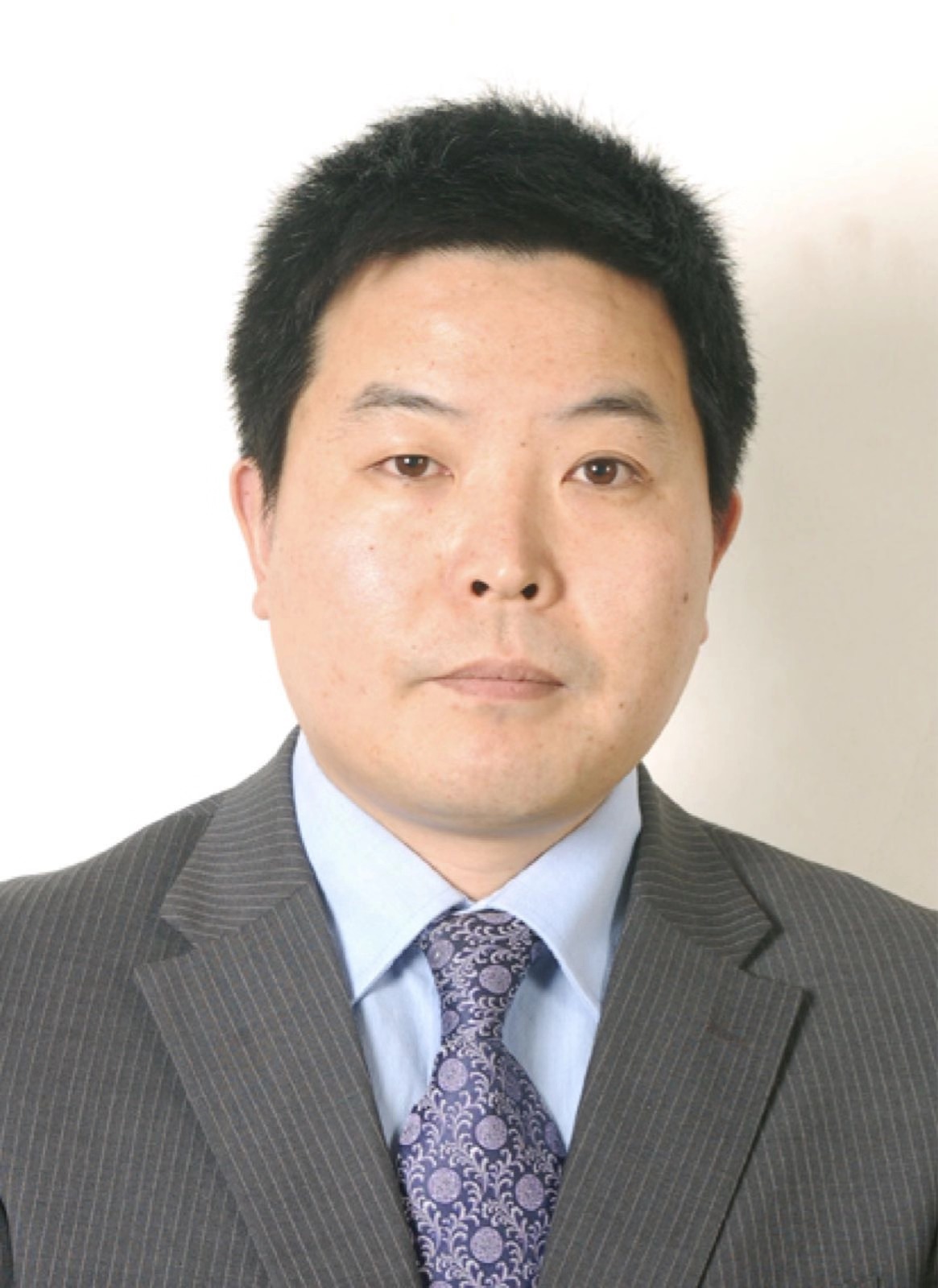}}]{Jiawan Zhang} received the M.Sc. and Ph.D. degrees in Computer Science from Tianjin University in 2001 and 2004, respectively. Currently, he is a Full Professor at the College of Intelligence and Computing, Tianjin University, Tianjn, China. He serve(d) for academic events including the general co-chair of ChinaVis (2015, 2016), PacificVis (2019, 2020). He also serve(d) as the program committee member or reviewer for many conferences and journals including CVPR, ICCV, AAAI, VIS, PacificVis, EuroVis, IEEE TVCG, IEEE TIP.
\end{IEEEbiography}






\end{document}